\documentclass[journal]{IEEEtran}
\usepackage[dvipsnames]{xcolor}
\usepackage{cite}
\usepackage[CJKbookmarks=true]{hyperref}

\hypersetup{hidelinks,
	colorlinks=true,
	allcolors=black,
	pdfstartview=Fit,
	breaklinks=true}

\newcommand{\reftab}[1]{Table \ref{#1}}

\ifCLASSINFOpdf
\usepackage[pdftex]{graphicx}
\newcommand{\reffig}[1]{Fig. \ref{#1}}
\graphicspath{{image}}
\DeclareGraphicsExtensions{.pdf,.jpeg,.png}
\else
\fi
\usepackage{amsmath}
\usepackage{amsthm}
\theoremstyle{plain}

\usepackage{amssymb}
\usepackage{multirow}
\usepackage{tabulary}
\usepackage[utf8]{inputenc}
\usepackage[switch]{lineno}
\usepackage{subfigure}
\begin{document}
\bstctlcite{IEEEexample:BSTcontrol}
\title{Rethinking Lightweight Convolutional Neural Networks for Efficient and High-quality Pavement Crack Detection}

\author{Kai~Li,~Jie~Yang,~Siwei Ma,~Bo~Wang, Shanshe~Wang, Yingjie~Tian,~and~Zhiquan~Qi

\thanks{This work was supported in part by the National Natural Science Foundation of China under Grant 12071458, Grant 71731009, and Grant 61702099; and in part by the Fundamental Research Funds for the Central Universities in UIBE Excellent Young Scholar Project (21YQ10).}

\thanks{Kai Li is with the School of Computer Science, Institute of Digital Media, Peking University, Beijing 100871, China (e-mail: kaili@pku.edu.cn).}

\thanks{Jie~Yang is with the School of Mathematics Sciences, University of Chinese Academy of Sciences, Beijing 100049, China (e-mail: yangjie181@mails.ucas.ac.cn).}

\thanks{Siwei Ma and Shanshe Wang are with the School of Computer Science, Institute of Digital Media, Peking University, Beijing 100871, China,
also with Advanced Institute of Information Technology, Peking University, Hangzhou 310005, China (e-mail: swma@pku.edu.cn and sswang@pku.edu).}


\thanks{Bo Wang is with the School of Information Technology and Management, University of International Business and Economics, Beijing 100029, China (e-mail: wangbo@uibe.edu.cn).}

\thanks{Yingjie Tian (Corresponding Author) and Zhiquan Qi are with the Research Center on Fictitious Economy and Data Science, Chinese Academy of Sciences, Beijing 100190, China; with the Key Laboratory of Big Data Mining and Knowledge Management, Chinese Academy of Sciences, Beijing 100190, China. Besides, Yingjie Tian is also with MOE Social Science Laboratory of Digital Economic Forecasts and Policy Simulation at University of Chinese Academy of Sciences, No. 3, South Yitiao, Zhongguancun, Haidian District, Beijing 100190, China.
(e-mail: tyj@ucas.ac.cn and qizhiquan@foxmail.com).}

}

\maketitle

\begin{abstract}

Pixel-level road crack detection has always been a challenging task in intelligent transportation systems.
Due to the external environments, such as weather, light, and other factors, pavement cracks often present low contrast, poor continuity, and different sizes in length and width.
{However, most of the existing studies pay less attention to crack data under different situations.}
Meanwhile, recent algorithms based on deep convolutional neural networks (DCNNs) have promoted the development of cutting-edge models for crack detection.
Nevertheless, they usually focus on complex models for good performance, but ignore detection efficiency in practical applications.
{In this article, to address the first issue, we collected two new databases (i.e. Rain365 and Sun520) captured in rainy and sunny days respectively, which enrich the data of the open source community.}
For the second issue, we reconsider how to improve detection efficiency with excellent performance, and then propose our lightweight encoder-decoder architecture termed CarNet.
{Specifically, we introduce a novel olive-shaped structure for the encoder network, a light-weight multi-scale block and a new up-sampling method in the decoder network.}
Numerous experiments show that our model can better balance detection performance and efficiency compared with previous models.
{Especially, on the Sun520 dataset, our CarNet significantly advances the state-of-the-art performance with ODS F-score from 0.488 to 0.514.
Meanwhile, it does so with an improved detection speed (104 frame per second) which is orders of magnitude faster than some recent DCNNs-based algorithms specially designed for crack detection.}
\end{abstract}


\IEEEpeerreviewmaketitle

\section{Introduction}

\begin{figure}[htbp]
    \centering
        \scriptsize
    \includegraphics[width=8.8cm]{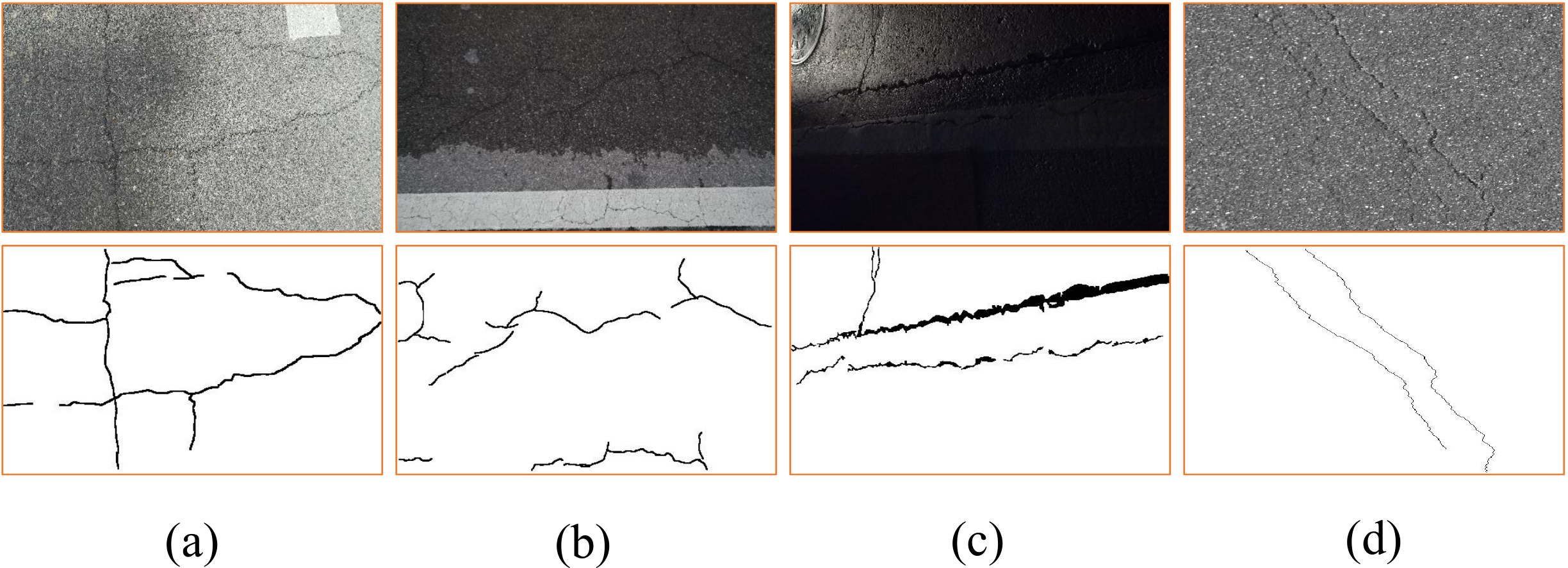}
    \caption{
    {Visual examples of interference factors.
    (a), (b), (c), and (d) contain shadows on sunny days, rainwater on rainy days, uneven illuminations in night scenes, and grain-like textures in the background images, respectively.
    The above images and their ground-truth originate from our Sun520, our Rain365, BJN260 \cite{li2023fast}, and Crack360~{\cite{zou2018deepcrack}}, respectively.}
    }
    \label{cracks with noise}
\end{figure}

\IEEEPARstart{C}{racks}, as common pavement defects, provide a sign of potential road damage.
Timely and accurate crack detection is very helpful to prevent damage expansion and ensure traffic safety.
In practice, pavement cracks usually exhibit low contrast, poor continuity, and different shapes and sizes, which are caused by intricate interference factors in \reffig{cracks with noise}.
Then, pixel-level crack detection remains a challenging problem so far.

{It is extremely rich for the history of road crack detection.
Many great achievements have been made in this field.
In terms of data, open-source pixel-level labeled crack databases have been continuously improved, ranging from dozens of gray-scale images in the early days to hundreds of color images now.
For example, AigleRN~\cite{2011Automatic} contains 38 gray-scale images while CFD \cite{shi2016automatic} has 118 RGB pictures.
Besides, the databases covers scenes from daytime to nighttime.
For instance, BJN260 \cite{li2023fast} is a road crack dataset for night scene.}

{Meanwhile, in terms of modeling, a series of related works have emerged since the advent of automatic crack detection methods.
We now highlight some representative works that have proven to be of great practical importance.
Specifically, one may categorize the existing methods into the following groups:
(1) early pioneering methods, such as wavelet transform \cite{subirats2006automation}, contrast ratio \cite{achanta2008salient}, texture-analysis \cite{hu2010novel}, and minimal path selection \cite{kaul2012detecting, amhaz2014new, Amhaz2016Automatic};
(2) traditional machine learning systems, such as CrackForest \cite{shi2016automatic} and some other appoaches \cite{oliveira2013automatic, strisciuglio2017detection, strisciuglio2019robust};
(3) methods based on deep learning, especially deep convolutional neural networks (DCNNs), such as \cite{zou2018deepcrack, liu2019fpcnet, yang2019feature, song2020automated, Li2020Automatic, Qu2021AMMFF, Qu2021MFF}.
Due to automatic feature extraction, recent DCNNs-based systems have shown promising F-score performance improvements over CrackForest.}

\begin{figure}[htbp]
    \centering
    \includegraphics[width=\linewidth]{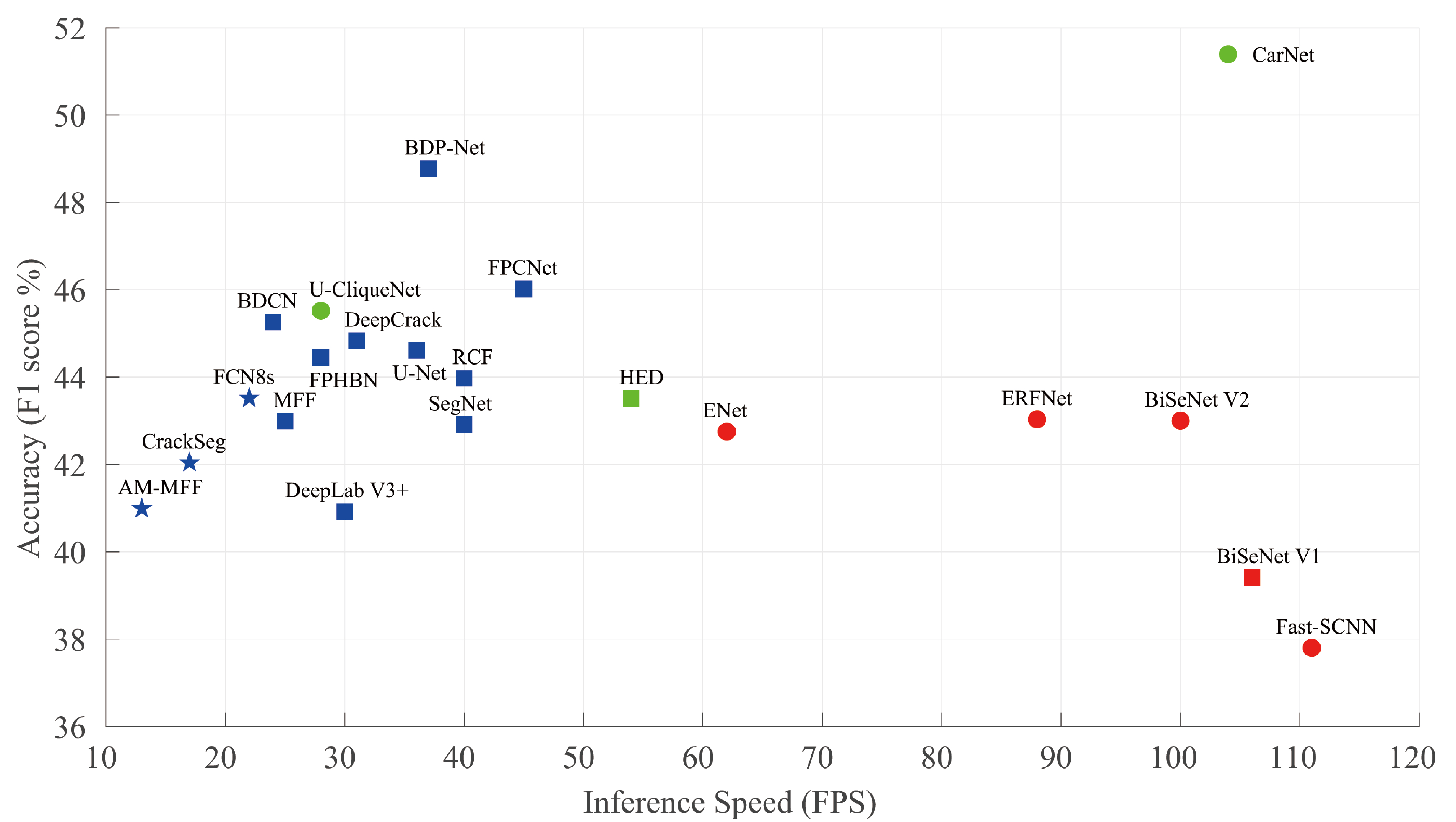}
    \caption{Model accuracy v.s. inference efficiency on Sun520 test set.
    The model accuracy is shown by the F$_1$ score of the whole test set.
    The inference efficiency is reflected by average frames per second (FPS) in a single NVIDIA 1080Ti.
    According to the model parameter intervals [0, 10), [10, 50), and [50, 300) in millions, we divide these methods into three categories represented by solid circle, square, and pentagram, respectively.
    Based on the floating-point operations, we utilize the above interval segments in gigabit to classify different systems into three groups corresponding to red, green, and blue, respectively.
    For more comparisons, we also show other methods \cite{xie2015holistically, liu2017richer, he2019bi, electronics2021likai,
    long2015fully, ronneberger2015u, badrinarayanan2017segnet, chen2018encoder,
    paszke2016enet, romera2018erfnet, yu2018bisenet, poudel2019fast, yu2021bisenet,
    zou2018deepcrack, liu2019fpcnet, yang2019feature, song2020automated, Li2020Automatic, Qu2021AMMFF, Qu2021MFF} for crack detection.
    }
    \label{speed_f1}
\end{figure}

{However, to our knowledge, the existing pavement crack databases are usually captured on clear weather.
The crack data under other weather, such as rainy day, are absent in this field.
Besides, the open-source single dataset for sunny days is small in data scale, e.g. it usually contains no more than 400 images.
Meanwhile, the state-of-the-art methods for crack detection tend to DCNN-based ones which always exchange large and heavy models for graceful accuracy.
In turn, bloated structure and complicated calculation also make them inefficient and inapplicable in practice, as shown in \reffig{speed_f1}.}

{In this paper, to address the above first issue, we collected two new pavement crack databases, namely Rain365 and Sun520, and annotated them at the pixel level.
Specifically, Rain365 contains 365 images captured after rain, covering pavement conditions with different degrees of dryness and wetness.
Sun520 includes 520 pictures taken from dawn to dusk on sunny days to illustrate varying light intensities.}
{For the second issue, we first think about why the existing crack detection methods based on DCNNs are bloated and inefficient, and find that encoder structures, multi-scale features, and feature up-sampling methods have non-negligible impacts on model performance and efficiency.
Then, based on three aspects, we propose corresponding improvement schemes which make up our framework termed \emph{CarNet}.
}

To sum up, our contributions can be summarized as follows:
\begin{itemize}
  \item
   Two pavement crack databases, i.e. Rain365 and Sun520, are established for performance evaluation and have been shared with the community to facilitate related research.
   As we know, in the open-source road crack datasets with pixel-level annotation, the former is the \emph{first} rain-scene one while the latter is the current \emph{largest} one.
  \item
    {In terms of modeling, based on encoder structures, multi-scale features and feature up-sampling ways, we analyze previous methods and propose a novel lightweight encoder-decoder framework for crack detection.}
  \item
    Sufficient experiments are conducted on various pavement crack databases for different scenarios.
    The experimental results show that our method achieves good efficiency/accuracy trade-offs over previous systems.
\end{itemize}

The remaining sections in this paper are organized as follows.
Section \ref{part2} briefly reviews related work.
Section \ref{part3} elaborate our method for crack detection.
Section \ref{part4} demonstrates experimentally the effectiveness of the proposed method.
Section \ref{part5} concludes this paper.
Besides, to verify the generality of our approach, we also reports performance comparisons for edge detection in Appendix \ref{Appendix C}.

\section{Related Work}\label{part2}


{In this section, we first specify DCNNs-based crack detection models, then state the existing feature up-sampling methods, and finally introduce the relations and differences between our method and existing systems.}

\subsection{{DCNNs-based Crack Detection Models}}

{According to the structure forms, we can simply divide DCNNs-based models into two categories: symmetric encoder-decoder architectures and asymmetric encoder-decoder frameworks.
If regarding road cracks as objects with certain shapes, one can utilize edge detection and image segmentation (abbreviated EDIS) algorithms for crack detection.
Accordingly, we also describe some recent EDIS-based models in this section.}

\paragraph{{Symmetrical encoder-decoder architectures}}
These models usually employ skip layers to merge the same resolution feature maps from their encoders and decoders, so as to compensate for the loss of spatial information in the down-sampling process.
For instance, U-Net~\cite{ronneberger2015u} took full use of various skip-layer to combine context and localization information from the contracting and expanding paths respectively.
FPCNet~\cite{liu2019fpcnet} constructed a fast U-shaped framework for crack detection by combining the multi-dilation module and the up-sampling module with the squeeze-and-excitation block~\cite{hu2018squeeze}.
U-CliqueNet~\cite{Li2020Automatic} separated cracks from background by integrating alternately updated cliques \cite{yang2018convolutional} into U-Net.
The above two models employed channel concatenation to fuse feature maps.
Different from them, SegNet~\cite{badrinarayanan2017segnet} transferred the max-pooling indices in the feature maps of the encoder to the same resolution feature maps in the decoder.

In addition, side-output networks further improve the multi-scale representation capability of U-shaped architectures for pavement crack detection.
DeepCrack~\cite{zou2018deepcrack} yielded hierarchical convolutional features by combing up-sampling and side-output technologies from SegNet~\cite{badrinarayanan2017segnet} and HED~\cite{xie2015holistically} respectively.
FPHBN~\cite{yang2019feature} applied side networks and hierarchical boosting  to carry out supervised learning and weight easy-and-hard samples respectively.
Besides, Qu et al.~\cite{Qu2021AMMFF} utilized the global context block~\cite{cao2019GCNet} and the side-output feature pyramid~\cite{xie2015holistically} as the attention mechanism and multi-features fusion respectively to develop an effective pavement crack detection algorithm.
To alleviate the gradual dilution of the high-level semantic information layer by layer, Qu et al.~\cite{Qu2021MFF} injected the highest level features directly into previous network stages and then built an effective deeply supervised model by the multi-scale feature fusion.
For convenience, we nominate the above two models as AM-MFF and MFF respectively in this paper.

Nonetheless, the symmetrical structure of the encoder and decoder often leads to excessive redundant parameters and calculations, and thus needs longer running time for inference.
Too many skip-layer and side-output networks reduce model parallelism and then cut down inference efficiency.

\paragraph{{Asymmetric encoder-decoder architectures}}
{According to the original processing scenes, we roughly divide these frameworks into two categories: ones for natural scenes and ones for street scenes.}

These models for natural scene images often apply one complex encoder (e.g. VGG~\cite{simonyan2015very} and ResNet~\cite{2016he}) to extract convolutional features, and then utilize one or more decoders to recover to the original input resolution.
For example, based on VGG~\cite{simonyan2015very} as an encoder network, fully convolutional networks (FCN) were first proposed in~\cite{long2015fully} to achieve end-to-end training for pixel-wise task.
Besides, some advanced methods based on edge detection algorithms (e.g. HED~\cite{xie2015holistically}, RCF~\cite{liu2017richer}, BDCN~\cite{he2019bi}, and BDP-Net~\cite{electronics2021likai}) applied a trimmed VGG as their encoders to extract features, and then restored feature maps of different stages directly to the original input resolutions to build their decoders.
Recent image segmentation-based approaches, such as DeepLab V3+~\cite{chen2018encoder} and CrackSeg~\cite{song2020automated}, employed a modified ResNet~\cite{2016he} as their encoder networks, and then fused high-level and low-level hierarchical features in their decoder neworks.
However, since directly adopting multi-class classification models for the large-scale database ImageNet~\cite{Deng2009ImageNet} as their encoders, these methods usually embody heavy parameter, calculations, and memory footprints.
Besides, excessive branch structures are not friendly to memory access, thereby cutting down model efficiency further.
As a result, most models end up having difficulty making effective inference.

The systems for street scene images can provide an effective mechanism for efficient crack detection.
For example, ENet~\cite{paszke2016enet} proposed that visual information was highly spatially redundant, and thus the input resolution needed be first compressed by the proposed early down-sampling strategy.
Apart from early down-sampling, ERFNet~\cite{romera2018erfnet} introduced a novel non-bottleneck module by combing a residual connection and factorized convolutions, so as to reduce parameters and calculations.
{In addition, BiSeNet series~\cite{yu2018bisenet, yu2021bisenet} presented bilateral segmentation networks, where the spatial and context paths were designed to preserve spatial details and semantic information respectively.
To our knowledge, BiSeNet V1 is the first real-time two-branch semantic segmentation network.
BiSeNet V2 further improved the model accuracy by adding some new strategies, such as guided aggregation layer and side output segmentation heads.}
To build a real-time pipeline, Fast-SCNN~\cite{poudel2019fast} incorporated a shared shallow network before spatial and context paths.
Nevertheless, due to the lack of multi-scale features of images, these models need to be further improved about model accuracy.

\subsection{{Feature Up-sampling Method}}

{For pixel-wise vision tasks, bilinear interpolation, un-pooling, and deconvolution become three common feature up-sampling methods.
Bilinear interpolation has been widely popularized due to its excellent efficiency.
However, this method often shows inferior model accuracy.
Un-pooling was introduced in SegNet~\cite{badrinarayanan2017segnet}, which could help achieve decent accuracy.
Nevertheless, it requires the decoder to have a similar structure to its encoder, especially on the convolutional channel, so as to facilitate the transfer of pooling indices.
In turn, un-pooling is accompanied by numerous parameters and heavy calculations.
Compared with the above two up-sampling methods, deconvolution introduced in~\cite{zeiler2014visualizing} is more popular due to its excellent flexibility and model accuracy.
Unfortunately, in order to improve model accuracy, the existing methods usually employ large kernel deconvolution which increases model complexity and reduces model efficiency potentially.
For example, HED~\cite{xie2015holistically} utilized deconvolutions with kernel size 2s when the up-sampling stride was s.
}

\subsection{{Relations and Differences with Existing Methods}}

{Motivated by asymmetric encoder-decoder frameworks, we introduce a novel light-weight architecture so as to achieve good efficiency/accuracy trade-offs.
Specifically, we draws on the encoder design of ENet~\cite{paszke2016enet} and ERFNet~\cite{romera2018erfnet}, i.e., using early down-sampling strategy to improve model efficiency.
Meanwhile, we learn from multi-scale feature design of HED~\cite{xie2015holistically} and DeepLab V3~\cite{chen2017rethinking} which merge hierarchical features from different network stages.}

{On the other hand, our model is different from previous models.
For encoding networks, the previous models (i.e. ENet and ERFNet) and our model belong to \emph{Type-1} and \emph{Type-2} methods in \reftab{different_type_encoders} of Section \ref{part3.1}, respectively.
The former tends to model acceleration and focus on reducing computational complexity.
The latter tends to model compression and focuses on reducing network parameters, which can save model storage space and facilitate deployment on end-side devices.
Last but not least, our CarNet with the residual block combinations (7, 6, 2) shows significant advantages in model performance while maintaining efficient inference.}

{For decoding networks, different from HED~\cite{xie2015holistically} and DeepLab V3~\cite{chen2017rethinking} using large kernel deconvolution and bilinear interpolation respectively for up-sampling, we apply small kernel deconvolution like 3$\times$3 with feature refinement block to trade off model performance and efficiency.
In addition, our multi-scale method differs slightly from previous multi-scale systems in terms of feature fusion mode and quantity, which are elaborated in Section \ref{Decoder}.
}

\section{Methodology}\label{part3}

In this paper, we regard crack detection as a pixel-wise binary classification problem.
Every pixel in images is divided into crack or non-crack.
To achieve a good efficiency/accuracy trade-off, our model is based on an asymmetric encoder-decoder framework rather than a U-shaped structure.

\begin{figure*}[htbp]
  \centering
  \includegraphics[width=\linewidth]{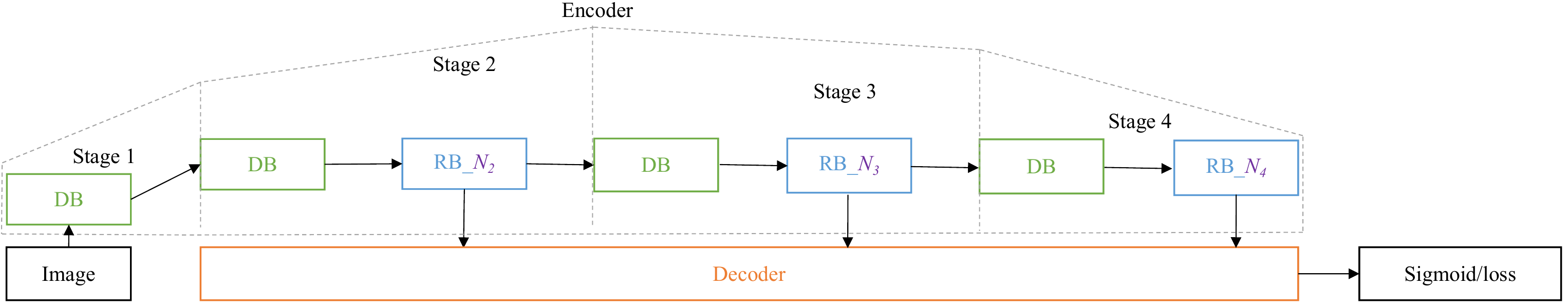}
  \caption{Our framework CarNet. Here, DB and RB refer to the down-sampling block and the residual block, respectively.
  $RB\_N_2$ means applying the residual block $N_2$ times in a cascaded mode.
  According to the design mechanism of our olive-shaped encoder, one may find that $N_2 \ge N_3 \ge N_4$, where $N_2, N_3$, and $N_4$ are the number of the residual blocks in the second, third, and fourth stages, respectively.
}
  \label{CarNet}
\end{figure*}

Based on the description of asymmetric encoder-decoder frameworks in Section \ref{part2}, we argue that encoder structures, multi-scale features, and feature up-sampling methods have great impacts on model performance and efficiency.
Then, we propose a novel olive-shaped structure for the encoder network, a light-weight multi-scale block and a new feature up-sampling way in the decoder network.
Finally, they make up our overall framework, as shown in \reffig{CarNet}.
For convenience, we nominate the proposed model \emph{CarNet} according to its shape.
Below, we introduce them in turn.

\subsection{The Proposed Olive-type Encoder}\label{part3.1}

\begin{figure}[htbp]
    \centering
    \subfigure[\tiny{}]{
    \centering
    \includegraphics[width=0.35\linewidth]{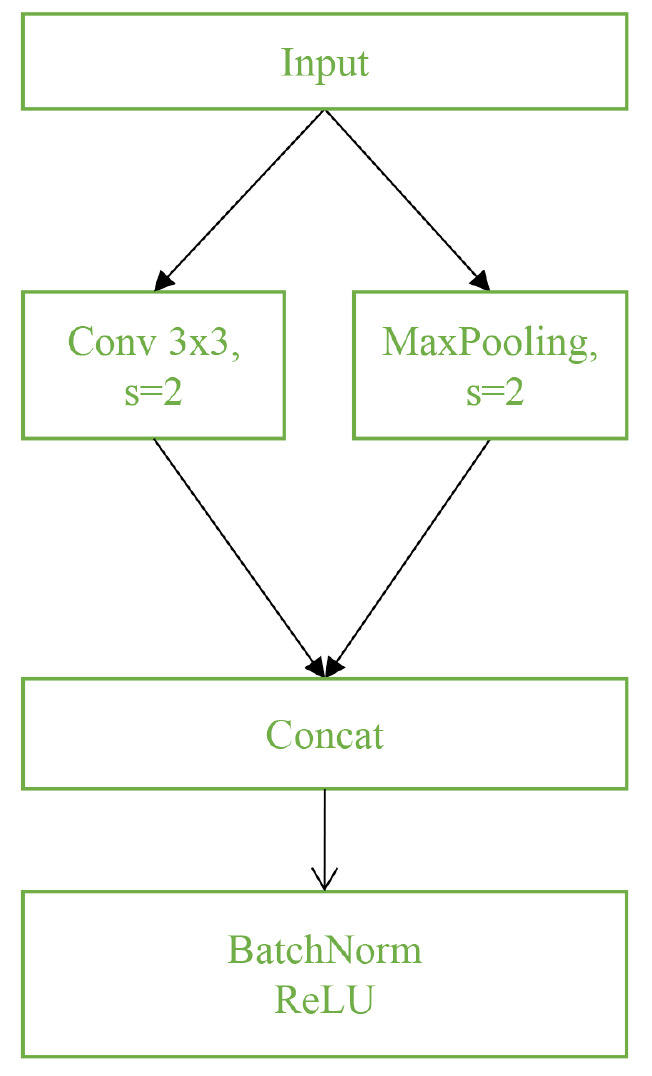}
    \label{CarNet_encoder_blocks:a}
    }
    \hspace{0.075\linewidth}
    \subfigure[\tiny{}]{
    \centering
    \includegraphics[width=0.4\linewidth]{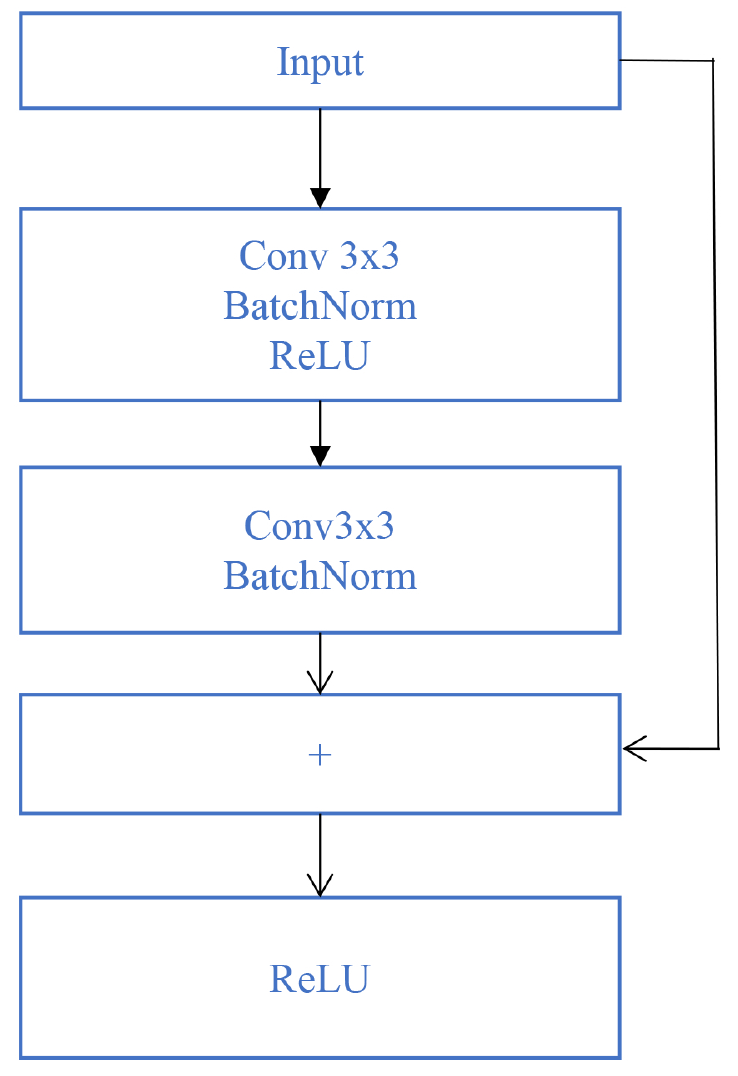}
    \label{CarNet_encoder_blocks:b}
    }
   \caption{Components in our encoder network.
   Here, Conv~$m \times n$ represents one convolution with kernel size m $\times$ n  and stride 1 by default.
   BatchNorm and ReLU refer to batch normalization and rectified linear unit, respectively.
   (a) The down-sampling block (DB).
   Here s=2 indicates that the convolution kernel stride is 2.
   Concat represents using channel concatenation for feature merging.
   (b) The residual block (RB).
   It utilizes identity mapping as a skip-layer to combine the input and output of two stacked convolutions.
  }
  \label{CarNet_encoder_blocks}
\end{figure}

\subsubsection{Motivation}
In general, model complexity includes space complexity and time complexity.
Space complexity can be roughly estimated by model parameters (Params) while time complexity can be reflected by floating-point operations (FLOPs). In practice, convolutional neural networks are often used for computer vision tasks including crack detection.
Taking one convolution with a bias term as an example, the corresponding space and time complexity are as follows:
\begin{align}
   \text{Params} &= (K_w * K_h * C_{in}) * C_{out} + C_{out}, \label{param}\\
   \text{FLOPs} &= [(K_w * K_h * C_{in}) * C_{out} + C_{out}] * W_{out} * H_{out}, \label{FLOPs}
\end{align}
{where Params, FLOPs, $K_w$, $K_h$, $C_{in}$, $C_{out}$, $W_{out}$, and $H_{out}$ stand for space complexity, time complexity, the width of the convolutional kernel, the height of the convolutional kernel, input channel, output channel, the width of the output feature map, and the height of the output feature map, respectively.}

As network stages deepen, the corresponding convolutional channels usually increase in multiples.
According to formula \eqref{param}, under the premise that the total number of convolutional layers remains unchanged in the encoder networks, deeper network stages need to contain fewer convolutional layers so as to reduce the overall spatial complexity.

On the other hand, in view of formula \eqref{FLOPs}, to cut down the overall time complexity of encoder networks, the initial and tail network stages need to embody fewer convolutional layers while the middle network stages can yield more convolutional layers.
Such an idea mainly takes into account: high feature resolution in the initial stage and large convolutional channels in the tail phase.

{Based on the two aspects, we apply a small number of strided convolutions to compress the input image resolution in the initial network stage, and next reduce the number of convolutional layers as the network stage deepens.}
Then, the whole encoder presents an olive-shaped structure about the number of convolutional layers in different network stages.
For convenience, we name it the olive-type encoder.

\subsubsection{Design Steps and Details}
As shown by the dashed box in \reffig{CarNet}, our encoder consists of two kinds of modules, i.e. the down-sampling block (DB) and the residual block (RB),
which are introduced in ENet~\cite{paszke2016enet} and ResNet~\cite{2016he} respectively.
In the following, we elaborate on the specific design of the encoder network.

First, we determine that the encoder network is a four-stage pipeline where the network stage is distinguished by feature map size.
{Specifically, the encoder networks for pixel-wise crack detection often down-sample images to 1/16 of the original input resolution, such as FPCNet~\cite{liu2019fpcnet} and MFF~\cite{Qu2021MFF}.
Motivated by ENet~\cite{paszke2016enet}, we adopt the early down-sampling strategy in the initial network stage, so as to improve model efficiency.
The strategy is conducted by two consecutive DB modules.
Then our encoder consists of features corresponding to 1/2, 1/4, 1/8, and 1/16 of the original input resolution.}

Note that the input in DB module, as shown in \reffig{CarNet_encoder_blocks:a}, is first respectively fed to one $3\times3$ convolutional layer and one maximum pooling layer with both stride 2 in a parallel mode.
Next, feature maps from two branches are fused by channel concatenation.
Finally, batch normalization \cite{ioffe2015batch} and ReLU \cite{he2015delving} are successively employed to do further processing.

Second, we consider the number of the convolutional layers in the intermediate and tail stages.
Based on the olive-type structure of encoders, we propose that the number of convolutional layers begins to decrease after the input size is compressed twice in succession.
In other words, it meets such requirements: N$_2$ $\ge$ N$_3$ $\ge$ N$_4$, where N$_2$, N$_3$, and N$_4$ are separately the number of the residual blocks in the second, third, and fourth stages, as shown in \reffig{CarNet}.

\subsubsection{{More Discussions on Olive-type Encoder}}

{According to the distribution of the number of convolution layers in different network stages, we divide olive-type encoders into two categories: Type\,\,1 and Type\,\,2.
Starting from the second network stage, the method Type 1 first increases and then decreases the number of convolution layers as network stage deepens.
Meawhile, the method Type 2 gradually reduces the number of convolutional layers with network phase deepening.
Note that the method Type 2 is our CarNet.}

{To demonstrate the effect of different olive-shaped encoders on model performance, we utilize slightly modified ResNet-34~\cite{2016he} as encoding networks and the decoder from \reffig{CarNet_decoder_blocks:c} as the decoding network to build different architectures.
Specifically, we first employ two residual blocks in the tail network stage\footnote{Note that for ResNet-34, besides a convolutional layer in the first stage and a fully connected layer at the end, the remaining network stages contains 3, 4, 6, and 3 residual modules, respectively.}, namely N$_4$ = 2.}
{Every residual block, as shown in \reffig{CarNet_encoder_blocks:b}, contains two stacked 3$\times$3 convolutional layers.
So the tail network stage contain 4 convolutional layers.
Besides, four down-sampling blocks also include 4 convolutional layers.
Then, the remaining 26 convolutional layers can be used in the intermediate network stages.
Specifically, in the second and third network stages of the encoder, the residual block combinations (N$_2$, N$_3$) like (3, 10), (4, 9), (5, 8), and (6, 7) can be employed for the method Type\,\,1 while the combinations (N$_2$, N$_3$) like (7, 6), (8, 5), (9, 4), and (10, 3) can be used for the method Type\,\,2.
}

\begin{table}[htbp]
    \centering
    \caption{Effect of convolutional layer distribution in encoder networks on model performance.
    We conduct these experiments on Sun520.}
    \resizebox{0.95\columnwidth}{!}{
    \begin{tabular}{c|c|c|c|c|c|ccc}
        \hline
            Method & PC & ODS & OIS & Params & FLOPs & FPS\\
        \hline
            \multirow{4}{*}{Type\,\,1}
                       & 3, 10, 2   & 0.4437  & 0.4533 & 5.78 M  & {11.25 G} & 99.63  \\
                       & 4, 9, 2    & 0.4450 & 0.4554 & 5.56 M & 11.25 G & 101.53  \\
                       & 5, 8, 2    & 0.4422 & 0.4555 & 5.34 M  & 11.25 G & 100.51  \\
                       & 6, 7, 2    & {0.4854}  & {0.4921} & 5.12 M  & 11.25 G & 102.41  \\
        \hline
            \multirow{4}{*}{Type\,\,2}
                        & 10, 3, 2   & 0.4396 & 0.4522 & \textbf{4.23 M}  & 11.26 G  & 97.36  \\
                        & 9, 4, 2    & 0.4440 & 0.4554 & 4.45 M  & 11.26 G  & 99.50  \\
                        & 8, 5, 2    & {0.4872} & {0.4874} & 4.67 M  & 11.26 G  & 95.83  \\
                        & 7, 6, 2    & \textbf{0.5139}  & \textbf{0.5158}  & {4.89 M}  & 11.26 G  & \textbf{104.37} \\
        \hline
    \end{tabular}
    }
    \label{different_type_encoders}
\end{table}

{As shown in \reftab{different_type_encoders}, compared with the models of Type 1, the proposed CarNet family display well-matched or better performance with slightly fewer parameters, slightly more calculations, and almost comparable inference efficiency.
For example, compared with the model using [6, 7, 2] residual block combinations in the encoder, the one with [7, 6, 2] combination gains at least 2\% increases in ODS and OIS respectively while getting well-matched parameters, calculations, and inference speed.
It shows that the performance of the encoder prefers but does not overindulge features from the second stage of the encoder network.
This further verifies the effectiveness of the proposed olive-type encoder.}


\subsubsection{{Design Innovation}}
{Previous state-of-the-art crack detection methods, such as DeepCrack~\cite{zou2018deepcrack} and MFF~\cite{Qu2021MFF}, mainly adopt existing multi-classification models as their encoders.
As the network stages deepen in these encoders, they often show an unabated trend for the number of convolutional layers in different stages.
Accordingly, these methods contain many redundant parameters and calculations.
To alleviate the above issue, we rethink the distribution of the number of convolutional layers in different network stages of the encoder, and propose a novel and light-weight olive-shaped encoder structure for pixel-wise crack detection.}

%

\subsection{The Proposed Lightweight Decoder} \label{Decoder}
Multi-scale features and feature up-sampling methods are two important factors that affect the performance and efficiency of the decoder network.
In this section, we rethink how to conduct them effectively and efficiently.

\subsubsection{{Multi-scale Features}}
In deep learning, the compression-and-fusion strategy is a common method to reduce model complexity and capture hierarchical features.
Motivated by this strategy in HED~\cite{xie2015holistically} and DeepLab V3+ \cite{chen2018encoder}, we introduce a lightweight multi-scale feature module, i.e. up-sampling feature pyramid block (UFPB).
Specifically, as shown in \reffig{CarNet_decoder_blocks:a}, three 1$\times$1 convolutions are employed to compress the last feature maps from the second, third and fourth stages respectively into the same feature channels.
Then, feature maps from the third and fourth stages are double and quadruple up-sampled respectively, so that their resolution is the same as that of the second stage.
Finally, one addition operation is used to merge features of different network stages.

\subsubsection{{Feature Up-sampling Method}}
Due to excellent flexibility and model accuracy, deconvolution introduced by~\cite{zeiler2014visualizing} becomes the most common up-sampling method for pixel-wise vision tasks.
The existing models usually employ large kernel deconvolution so as to improve model accuracy.
For instance, HED~\cite{xie2015holistically} applied deconvolutions with kernel size 2s when the up-sampling stride was s.
However, large kernel potentially increases model complexity and reduces model efficiency.

\begin{figure}[htbp]
  \centering
    \subfigure[\tiny{}]{
        \centering
        \includegraphics[width=0.75\linewidth]{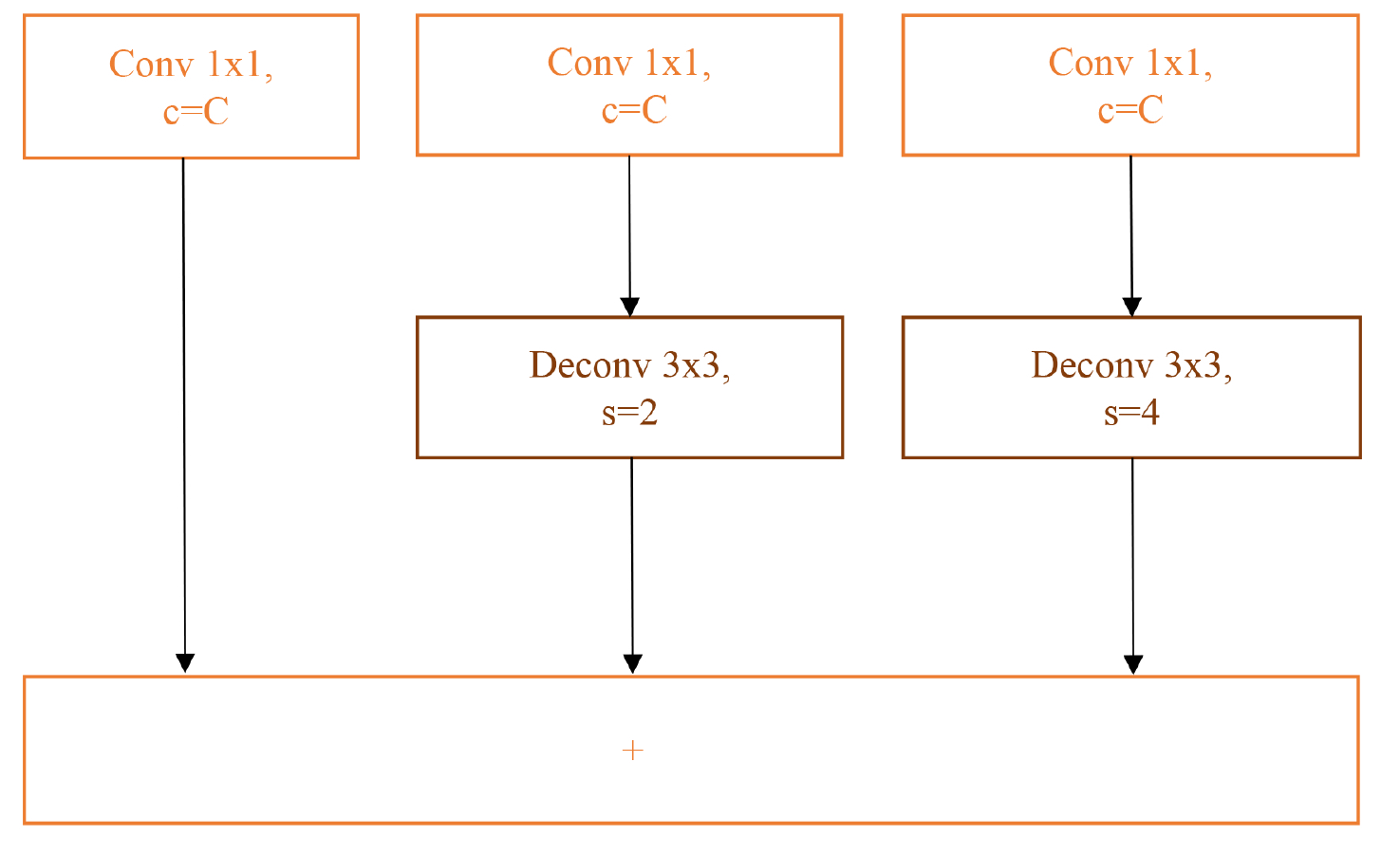}
        \label{CarNet_decoder_blocks:a} %
    }
    \subfigure[\tiny{}]{
        \centering
        \includegraphics[width=0.28\linewidth]{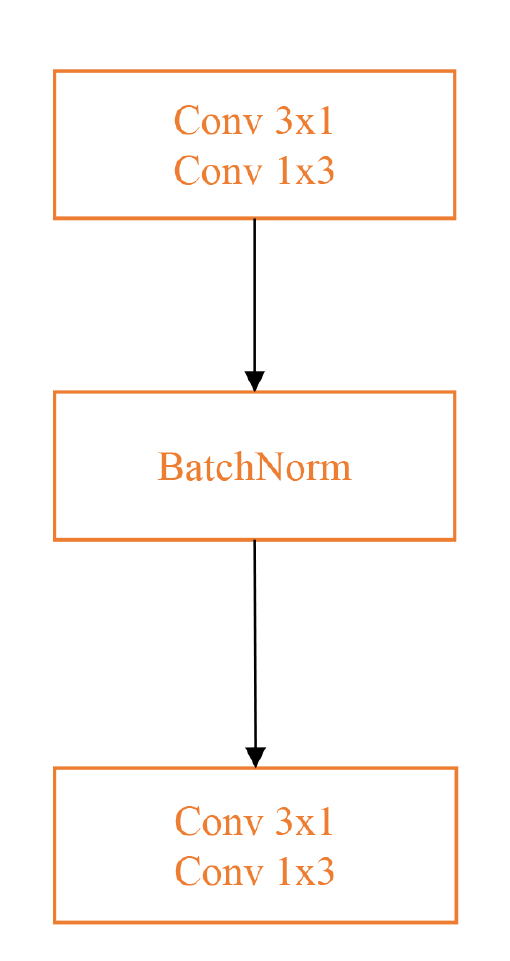}
        \label{CarNet_decoder_blocks:b} %
    }
     \hspace{0.15\linewidth}
    \subfigure[\tiny{}]{
        \centering
        \includegraphics[width=0.285\linewidth]{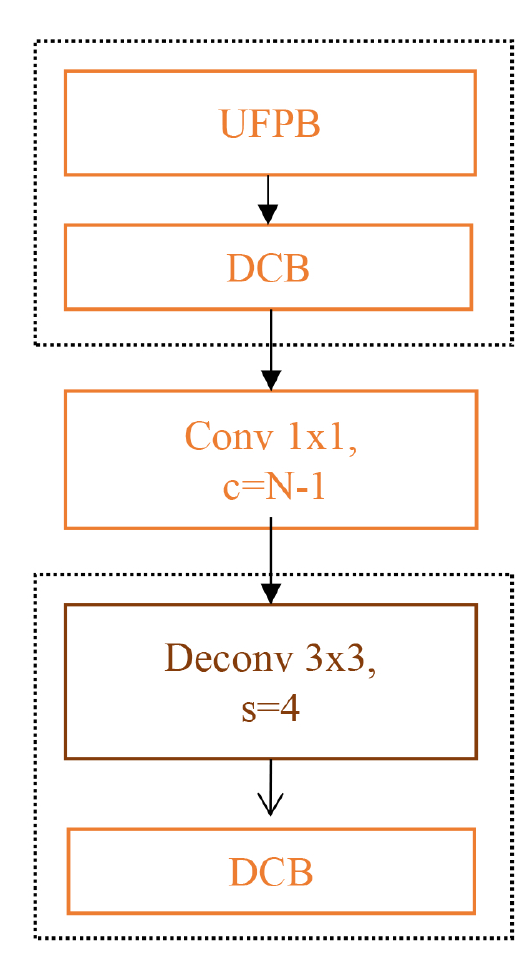}
        \label{CarNet_decoder_blocks:c} %
    }
      \caption{Component and framework of our decoder.
        (a) Our up-sampling feature pyramid block (UFPB).
            Conv 1$\times$1, c=C indicates using 1$\times$1 convolution to compress feature channel to C.
            Deconv means using deconvolution for up-sampling.
            Here we employ three 1$\times$1 convolutions to compress the last features from the last three network stages of our encoder respectively, and then merge three features by one addition operation.
        (b) Our decomposition convolution block (DCB).
        It contains two pairs of cascaded 3$\times$1 and 1$\times$3 convolutions with one batch normalization in the middle.
        (c) Our decoder framework.
            It can be seen as consisting of the contents contained by the dotted boxes at both ends and the 1$\times$1 convolution in the middle.
            Here N represents the number of categories, such as N=2 for crack detection.
      }
  \label{CarNet_decoder_blocks}
\end{figure}

\begin{figure*}[htbp]
  \centering
  \scriptsize
  \includegraphics[width=\linewidth]{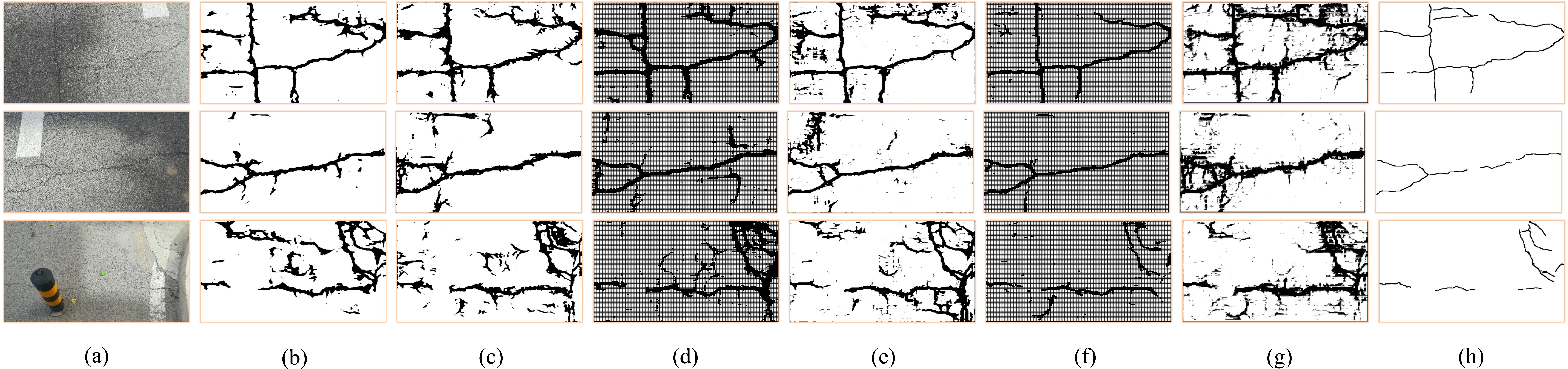}
  \caption{{Visualization of different up-sampling methods.
  (a) and (h) are three images randomly selected from the Sun520 test set and the corresponding ground truth images, respectively.
  (b$\sim$d) show the predicted images obtained by combing our CarNet and various up-sampling methods (i.e. bilinear interpolation, large kernel deconvolution, and 3$\times$3 small kernel deconvolution) without any DCB module, respectively.
  (e$\sim$f) correspond to the results obtained by combing our CarNet and 3$\times$3 small kernel deconvolution without first DCB,  without second DCB, with two DCB modeules, respectively.
  In order to observe gridding effects clearly, we recommend zooming in on the above images.
  }
  }
  \label{Visualization of different up-sampling methods}
\end{figure*}

Since VGG family \cite{simonyan2015very} emerged, small kernel convolution like 3$\times$3 has become widely popular in deep learning.
Why not use small kernel deconvolution instead of large kernel deconvolution?
This is because directly using small kernel deconvolution for feature up-sampling is prone to gridding effects on the test images, {as shown in \reffig{Visualization of different up-sampling methods} (d).}


To solve the gridding effect caused by small kernel deconvolution, we propose to refine up-sampled features by feature refinement module.
Since cracks are mostly linear structures, we perform feature refinement by a decomposition convolution block (DCB) in the decoder.
As shown in \reffig{CarNet_decoder_blocks:b}, this block embraces two pairs of cascaded decomposition convolutions (i.e. one $3\times1$ and another $1\times3$).

{In order to find the specific link that produces the gridding effect, we conduct further investigation experiments by reducing the module DCB.
By \reffig{Visualization of different up-sampling methods} (e) and \reffig{Visualization of different up-sampling methods} (f), one find that the gridding effect is caused by the last small kernel deconvolution.
Note that although first DCB module plays a small role in the de-grid effect, it helps to reduce the omission of cracks, as shown by the two images in the first row of \reffig{Visualization of different up-sampling methods} (e) and \reffig{Visualization of different up-sampling methods} (g).
For strict quantitative comparison results, please refer to \reffig{Upsampling_FR}.
}

\subsubsection{Design Steps and Details}
As shown in \reffig{CarNet_decoder_blocks:c}, in our decoder, the last features from the last three network stages of our encoder are fed successively to the multi-scale module UFPB and the feature refinement module DCB.
Then, feature is compressed to N-1 channels by 1$\times$1 convolution, where N refers to the number of categories.
Next, we utilize 3$\times$3 deconvolution to up-sample the compressed feature to the original input size, and employ the module DCB again to refine up-sampled feature.

To sum up, our decoder consists of three parts, i.e. UFPB followed with DCB, 1$\times$1 convolution, and small kernel deconvolution followed with DCB.
Finally, as shown in \reffig{CarNet}, feature from our decoder is fed successively to sigmoid classifier and loss function.

\subsubsection{Comparison between Our and Other Decoders}

\begin{figure*}[htbp]
  \centering
  \scriptsize
  \includegraphics[width=\linewidth]{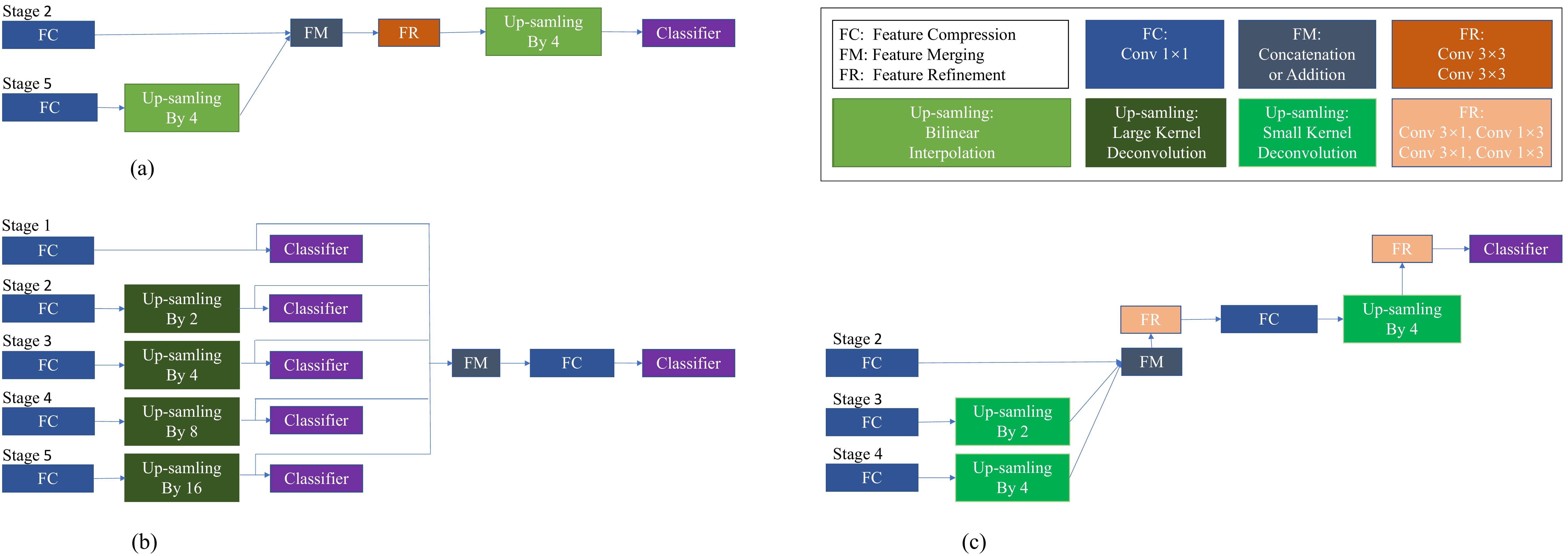}
  \caption{Decoder structure of different models.
  (a) DeepLab V3+ Decoder.
  (b) HED Decoder.
  (c) Our Decoder.
  Here, FC, FM, and FR refer to feature compression, feature merge, and feature refinement, respectively.
  For simplicity, we omit batch normalization used after convolutional layers.
  }
  \label{different_decoders}
\end{figure*}

Although our decoder is based on the ones from HED and DeepLab V3+, it is different from them in several respects, as shown in \reffig{different_decoders}.

\paragraph{Feature Up-sampling}
DeepLab V3+ applied bilinear interpolation for up-sampling to improve inference efficiency.
Meanwhile, HED employed large kernel deconvolution to enhance its performance.
Different from them, we utilize small kernel deconvolution with feature refinement to take into account model accuracy and efficiency.

\paragraph{Feature Fusion}
DeepLab V3+ employed features from the second and last stages in its encoder, which leaded to a lack of middle-level features.
HED utilized features from all network stages to construct an integrated feature pyramid network. Despite obtaining rich hierarchical information, it also involved excessive calculations.
Different from them, our model adopts features from the last three network stages to build a lightweight feature pyramid, to achieve a better trade-off between model performance and efficiency.

Besides, unlike DeepLab V3+ and HED using channel concatenation, our model adopts an addition operation for feature merging, so as to reduce model complexity further.

\paragraph{Decoder Branch}
Our model and DeepLab V3+ both employed a single decoder while HED utilized six ones corresponding to five side outputs and one fusion output respectively.
Thus, HED needed to weigh the loss functions for different outputs, which brings more challenges to model training.
Moreover, multi-branch output increases the memory access cost, which reduces model efficiency in turn.

\subsubsection{{Design Innovation}}

{In the decoder network, to avoid the gridding effect caused by 3$\times$3 small kernel deconvolution for feature up-sampling, we propose to combine small kernel deconvolution with the proposed feature refinement module DCB.
Meanwhile, compared with large kernel deconvolution, our up-sampling strategy can achieve better model accuracy and efficiency, as shown in \reffig{Upsampling_FR} and Footnote 4.}

\subsection{Loss Function}
Pixel-wise crack detection boils down to a binary classification problem.
Then we train the proposed model by the classical binary classification loss, i.e. cross-entropy:
\begin{equation}\label{eq}
    \small
    L(\mathbf{y}, \mathbf{p})=-\sum_{j}\big[ y_j \, log \, p_j + { (1-y_j) \, log(1-p_j)} \big],
\end{equation}
where $y_j$ and $p_j$ stand for the real label and the posterior possibility of the $j$-th pixel.
Note that $y_j = 1 $ and $y_j = 0 $ correspond to crack pixel and non-crack pixel, respectively.

\section{Experiments}\label{part4}
Here we start with benchmark databases and implementation details for crack detection.
Then, we elaborate on the effectiveness of the proposed architecture with ablation experiments.
Next, we report the performance comparison between our method and other state-of-the-art systems on various datasets.
Finally, we discuss model efficiency of different methods in detail.

\subsection{Experimental Datasets}
We evaluate various methods on three databases, namely our Sun520, our Rain365 and BJN260 \cite{li2023fast}.
We captured all images\footnote{{To sample diverse crack data, we selected different areas in Haidian District, Beijing, China.
Specifically, we collected the road surface conditions of the Zhongguancun Campus of University of Chinese Academy of Sciences, the auxiliary road of Zhongguancun East Road, Zhongguancun South Third Street, the Basic Science Park of Chinese Academy of Sciences, and the side road of North Fourth Ring West Road.}} using a mobile device Honor 6X and then labeled them pixel-by-pixel using {Image Labeler} in MATLAB.

\subsubsection{{Sun520}}
It consists of 520 images captured on sunny days.
These pictures contain various background noises, such as shadows, oil stains, lane marking, and curved lines of tiles.
Besides, since data were collected in the morning, afternoon, and dusk, Sun520 is rich in image brightness.

\subsubsection{{Rain365}}
It contains 365 images taken after rain.
Since human eyes hardly distinguish cracks and non-cracks in rainy night scenes, we only utilize post-rain images in the daytime.
Rain365 embraces three backgrounds, i.e., completely wet, partially wet and partially dry, and completely dry.
Among them, the first case is the majority.

\subsubsection{{BJN260}}
It includes 260 images captured under the night scene of Beijing.
Due to various light sources (e.g. street lamps, car lights, truck lights, etc) and different light intensities, pavement conditions at night are complex and changeable, which brings new challenges to pixel-level detection.

To save computing resources, we resize the image resolution in the three databases from $3968\times2240$ to $480\times320$.
For the division of Sun520, Rain365, and BJN260 datasets, we randomly select 400, 300, and 200 images for training respectively, and the remaining 120, 65, and 60 images for test respectively.

\subsection{Implementation Details}

\subsubsection{Experimental Environment}
All experiments are conducted on an NVIDIA GeForce GTX 1080Ti with AMD 3700X 8-Core Processor. We implement our network using the publicly available PyTorch 1.6~\cite{paszke2017automatic}.

\subsubsection{Data Augmentation}

In this paper, we utilize some data augmentation methods to increase the size of these crack databases.
Specifically, we first rotate the training images by 180 degrees, and next horizontally flip the original and rotated images.
Besides, we add noise to the original images by applying zero-mean Gaussian distributions with standard deviation 0.01.
Then, the augmented training set is four times larger than the original one.

\subsubsection{Hyper-parameters Settings}
Unless otherwise specified, the full images are employed as the model input.
The Adam optimizer \cite{kingma2015adam} is adopted to update network parameters with a mini-batch size of 2 in every iteration.
The initial learning rate is set to $3\times10^{-4}$.
Besides, feature maps are compressed to 32 channels in our multi-scale block.
The random seed is set to a constant 7, so as to alleviate random errors in the experiments.

{Unlike natural images, especially ones from ImageNet \cite{2009ImageNet}, crack images always contain intricate interference factors, such as impulse noises.
Besides, DeepCrack~\cite{zou2018deepcrack} experimentally demonstrated that the model trained from scratch can obtain better performance than that trained from pre-trained weights.
Thus, we train all the crack detection models from scratch instead of pre-trained weights from ImageNet.}
In experiments, we save and test our model every five training epochs for optimal performance.
Especially, our CarNet obtains the best experimental results with 10, 15, 25, and 30 training epochs on Sun520, Rain365, BJN260, and Crack360, respectively.

\subsubsection{Evaluation Metrics}
In this paper, we prefer to employ F$_1$ score defined as
\begin{align}\label{F1 score}
  \textbf{F${_1}$} = \frac{2 \times precision \times recall}{precision + recall},
\end{align}
for the evaluation of model performance,
which considers the limitations of precision and recall for the imbalance problem of crack and background pixels.

When evaluating the similarity between the prediction image and ground-truth, it refers to two different threshold methods,
i.e., optimal dataset scale (ODS) and optimal image scale (OIS),
where the optimal threshold is based on the whole dataset and every single image, respectively.
In this work, we adopt ODS and OIS as representatives to report two corresponding F$_1$ scores for convenience.

In addition, we also present space and time complexity by parameters (Params) and floating-point operations (FLOPs) respectively for model evaluation.
Meanwhile, we report inference efficiency using average frames per second (FPS).

\subsection{Ablation Study}\label{effect of network}

In this section, we first introduce the benchmark model for crack detection and then demonstrate the validity of the proposed encoder and decoder networks respectively.

\paragraph*{1) Baseline Model}{\label{baseline}}
For pixel-level tasks including crack detection, common methods utilize ResNet~\cite{2016he} as encoders to combine one or more decoders.
Given the limited scale of crack data, we choose ResNet-34 with minor revisions as the benchmark encoder.
Compared with the original model, we remove the average pooling layer and the fully connected layer in the trimmed ResNet-34, so as to reduce the model computation and storage workload.

For the decoder network, we gradually merge feature maps from the fifth and fourth stages, the fourth and third stages, and the third and second stages.
During this process, $3 \times 3$ deconvolutions are used to compress feature channels and increase image resolution, and then one addition is conducted for feature fusion.
Next, the fused feature maps are restored to the model input sizes by quadruple up-sampling.
Finally, for fair comparisons with our CarNet, the feature refinement module DCB is also used after the above feature fusion.

\paragraph*{2) The Validity of The Proposed Encoder and Decoder}
\reftab{permutation_and_combination} shows the relevant experimental results\footnote{
Here $\textbf{E}\_C\_\textbf{D}\_B^{\ddag}$ refers that the encoder network is derived from CarNet, and the decoder network like Baseline$^{\ddag}$ gradually merges features from different network stages.
Besides, $\textbf{D}\_cB^{\ddag}$ indicates that the decoder performs the compress-and-upsampling strategy like the decoder of our CarNet and then gradually conducts feature merging like the decoder of Baseline$^{\ddag}$.
For simplicity, we omit all the models’ initial network stages which are used to compress the input image resolution.
}.
Note that Baseline$^{\dag}$ and Baseline$^{\ddag}$ are the benchmark models.
The encoder of Baseline$^{\dag}$ applies the same output channel combinations with the original ResNet-34, i.e. $[64, 128, 256, 512]$.
Meanwhile, the encoders of other models including Baseline$^{\ddag}$ employ the same output channels,
i.e. $[16, 64, 128, 256]$, to reduce model parameters and calculations.

\begin{table}[htbp]
    \centering
    \small
    \caption{Ablation experiments on Sun520.
    {The letters \textbf{E} and \textbf{D} refer to the encoder and decoder, respectively.
    C and B$^{\ddag}$ indicate CarNet and Baseline$^{\ddag}$, respectively.
    {$\star$} indicates that the corresponding model utilizes one 3$\times$3 convolution instead of 1$\times$3 convolution and 3$\times$1 convolution in the block DCB.}
    Besides, {PC} refers to the permutations and combinations of residual blocks in different network stages of the encoder.}
    \resizebox{0.998\columnwidth}{!}{
    \begin{tabular}{c|c|c|c|c|c|ccc}
        \hline
            Models & PC & ODS & OIS & Params & FLOPs & FPS\\
        \hline
         Baseline$^{\dag}$     & 3, 4, 6, 3 & 0.4391 & 0.4492   & 24.38 M & 14.89 G & 94.44  \\
         Baseline$^{\ddag}$    & 3, 4, 6, 3 & 0.4241 & 0.4434   & 6.42 M & \textbf{3.13 G}  & \textbf{107.92}  \\
        \hline
            \textbf{E}\_C\_\textbf{D}\_B$^{\ddag}$               & 7, 6, 2         & 0.4511    & 0.4619    & 5.46~M     & 13.38~G     & 101.11  \\
            \textbf{E}\_C\_\textbf{D}\_cB$^{\ddag}$              & 7, 6, 2         & 0.4450    & 0.4541    & 4.91~M     & 11.22~G     & 101.41   \\
            {CarNet$^{\ast}$}                    & {7, 6, 2}    & {0.4787}          & {0.4910}    &{4.9~M}    & {11.32~G}     & {105.08}   \\
            CarNet                                               & 7, 6, 2    & \textbf{0.5139}  & \textbf{0.5158}  & \textbf{4.89~M}  & 11.26~G  & {104.37} \\
        \hline
    \end{tabular}
    }
    \label{permutation_and_combination}
\end{table}


\begin{figure*}[htbp]
    \centering
    \includegraphics[width=0.491\linewidth]{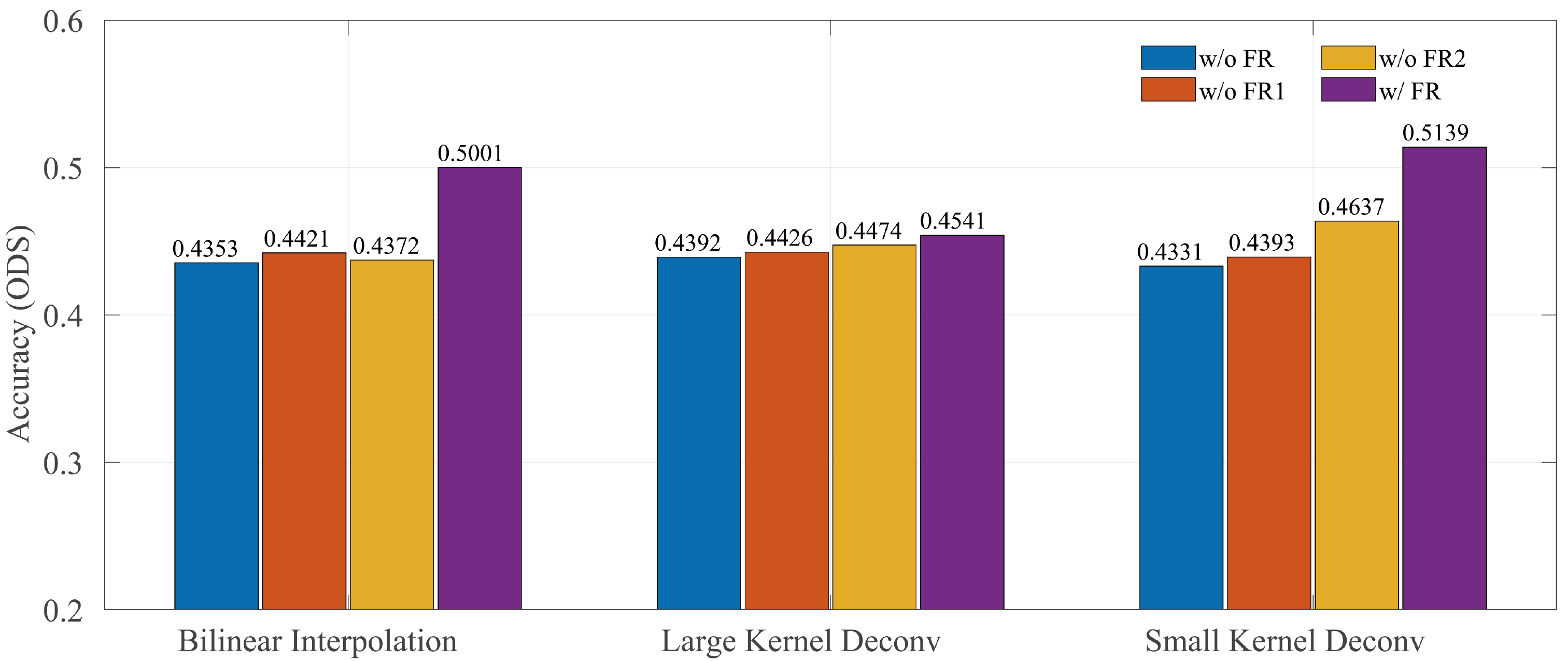}
    \hspace{0.005\linewidth}
    \includegraphics[width=0.491\linewidth]{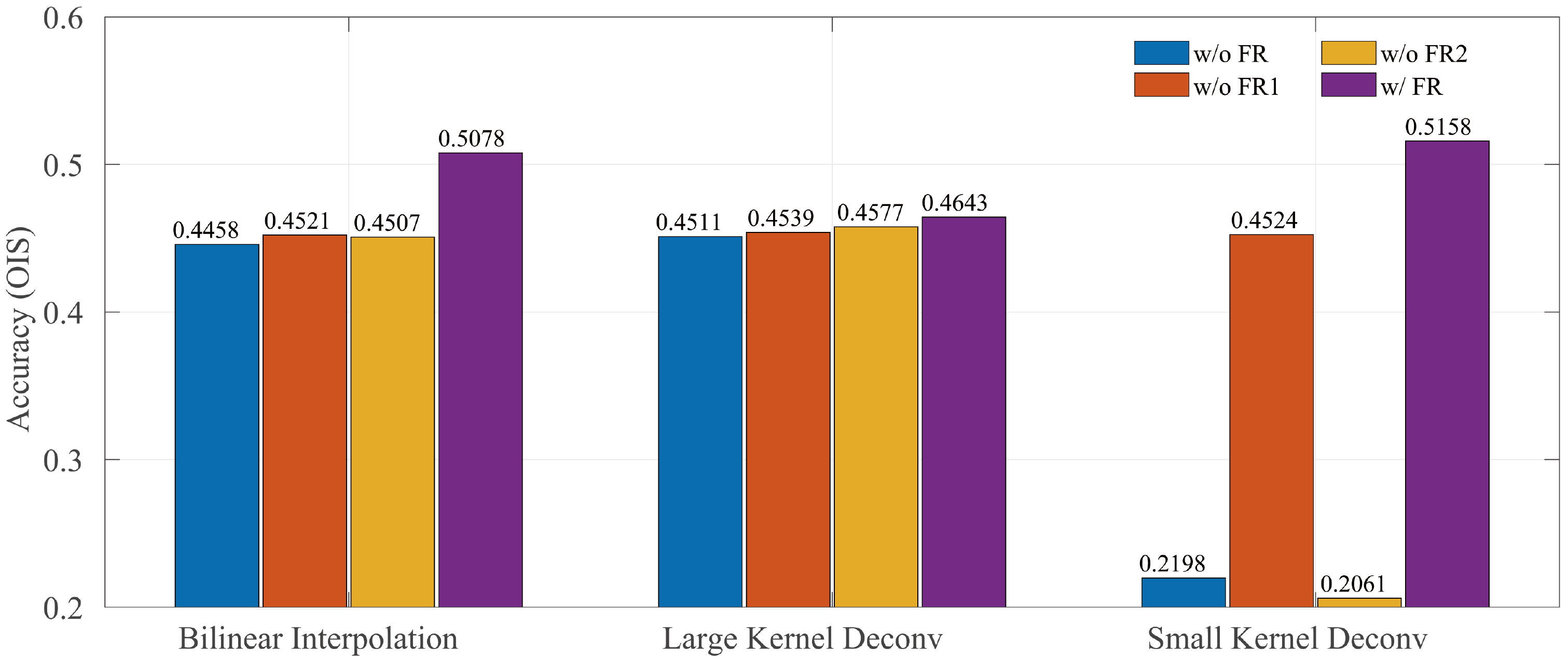}
    \caption{Model accuracy v.s. up-sampling method.
    Here the model accuracy ODS and OIS correspond to two $F_1$  scores based on different threshold methods.
    The up-sampling method involves three specific ways, i.e., bilinear interpolation, large kernel deconvolution, and small kernel deconvolution.
    FR stands for the feature refinement module DCB.
    Besides, w/o FR, w/o FR1, and w/o FR2 indicate the corresponding models without any DCB, first DCB, second DCB while w/ FR refers to the one using two DCB modules in the decoder network of \reffig{CarNet_decoder_blocks:c}.
    }
    \label{Upsampling_FR}
\end{figure*}

\textbf{The Proposed Olive-type Encoder:}
Note that Baseline$^{\ddag}$ and $\textbf{E}\_C\_\textbf{D}\_B^{\ddag}$ utilize similar decoder structure.
Compared with Baseline$^{\ddag}$ with [3, 4, 6, 3] residual block combinations in the encoder, $\textbf{E}\_C\_\textbf{D}\_B^{\ddag}$ with [7, 6, 2] residual block combinations obtains about 2\% performance gains in metrics ODS and OIS respectively while saving about 1~M model parameters.
Although in comparison with Baseline$^{\ddag}$, ${\textbf{E}\_C\_\textbf{D}\_B^{\ddag}}$ increases about 10~Gigabit FLOPs, but its inference speed only drops by about 7 frames per second.
It indicates that the proposed olive-type encoder can achieve a good tradeoff between model performance and inference efficiency.

Meanwhile, compared to Baseline$^{\dag}$ with [3, 4, 6, 3] residual block combinations in the encoder, ${\textbf{E}\_C\_\textbf{D}\_B^{\ddag}}$ with [7, 6, 2] residual block combinations is nearly 7 frames per second faster about inference speed while gaining about 1\% advantages in ODS and OIS. Besides, our model saves about 19~M and 2~Gigabit in parameters and calculations, respectively.
Therefore, the proposed olive-shaped encoder is more lightweight, more efficient, and more effective than the method of enlarging feature channels in the encoder network.

\textbf{The Proposed Lightweight Decoder:}
Despite using the same residual block combinations [7, 6, 2] in the encoder, compared with ${\textbf{E}\_C\_\textbf{D}\_B}^{\ddag}$,
our CarNet has advantages in model performance, inference speed, and model complexity.
To be specific, our model yields about 6\% and 5\% gains in ODS and OIS respectively and 3 FPS improvement in inference speed while saving about 0.6 M and 2.1 Gigabit FLOPs in parameters and calculations respectively.
Accordingly, this reveals the effectiveness of the proposed decoder.

Note that ${\textbf{E}\_C\_\textbf{D}\_cB^{\ddag}}$ and ${\textbf{E}\_C\_\textbf{D}\_B^{\ddag}}$ have a slight difference in merging features from the previous network stage.
The former compresses feature channels and then resizes feature sizes, while the latter utilizes deconvolution to adjust them concurrently.
Compared with the former, the latter achieves better performance in terms of metric ODS and OIS.
It implicitly reveals that the performance difference between ${\textbf{E}\_C\_\textbf{D}\_B^{\ddag}}$ and our CarNet is not due to the large model complexity, but the reasonable design of the decoder.

Next, we demonstrate the validity of the proposed multi-scale block UFPB.
Note that ${\textbf{E}\_C\_\textbf{D}\_cB^{\ddag}}$ and our CarNet utilize the same encoder network and employ the compression-and-upsampling strategy in their decoders.
The difference between the two lies in the multi-scale features of the decoder.
Specifically, the former constructs multi-scale information by merging features from adjacent network stages while the latter employs the multi-scale module UFPB.
According to \reftab{permutation_and_combination}, compared with ${\textbf{E}\_C\_\textbf{D}\_cB^{\ddag}}$, our CarNet improves model accuracy at least 6\%.
Meanwhile, they both are almost equivalent in model complexity and inference efficiency.

{Moreover, as shown in \reftab{permutation_and_combination}, compared with CarNet$^{\ast}$, our CarNet improves ODS and OIS by at least 2\% and achieves similar inference efficiency.
It shows that the proposed DCB is more suitable for the feature refinement module than two cascaded 3$\times$3 convolutions.}
Besides, we experimentally illustrate how feature up-sampling method and feature refinement affect model performance.
According to the results shown in \reffig{Upsampling_FR}, we can draw the following conclusions:

a) With the feature refinement module, small kernel deconvolutions outperforms the other two feature up-sampling methods in model performance.
Specifically, it improves ODS and OIS by about 1\% over bilinear interpolation, respectively.
Meanwhile, it is at least 5\% higher than large kernel deconvolution in terms of ODS and OIS.
Besides, small kernel deconvolution is also superior to bilinear interpolation and large kernel deconvolution in terms of model efficiency\footnote{
Specifically, for the models that combine different up-sampling methods (i.e.bilinear interpolation, large kernel deconvolution, and small kernel deconvolution) with two DCB modules, their model parameters, floating-point operations, and inference speed are [4.88 M, 11.08 G, 102.80 FPS], [4.96 M, 11.87 G, 96.51 FPS], and [4.89 M, 11.26 G, 104.37 FPS], respectively.
}.

b) {For various up-sampling methods, the two feature refinement modules have various impact on the model performance.}
Specifically, for bilinear interpolation and small kernel deconvolution, they help achieve at least 6\% gains in model metrics ODS and OIS.
However, for large kernel deconvolution, they only help improve about 1.5\% and 1.3\% in ODS and OIS respectively.
We speculate that this may be because even combined with feature refinement, large deconvolution kernels still cannot fully utilize local features.

Besides, the two feature refinement modules have little impact on model complexity.
Specifically, for parameters and floating-point operations, the first feature refinement module increases about 0.02~M and 0.12~G respectively while the second one adds less than 0.01~M and about 0.11~G.

c) {The single feature refinement has different effects on different up-sampling methods.}
In terms of ODS and OIS, the second feature refinement module DCB has a slightly larger effect on bilinear interpolation.
For large kernel deconvolution, this conclusion is on the contrary.

On the other hand, for small kernel deconvolution, the first DCB has a stronger impact on ODS, while the second DCB plays a more important role in OIS.
{We attribute this phenomenon to the fact that the optimal threshold evaluation method based on optimal image scale are more susceptible to gridding effects when compared to the one based on optimal dataset scale.
Note that small kernel deconvolution without the second DCB is prone to gridding effects when up-sampling, as shown in \reffig{Visualization of different up-sampling methods}.
}


\subsection{Experimental Results on Various Databases} \label{Performance on Various Databases}

In this section, we compare the experimental results of different models on various crack databases.
According to the vision tasks initially handled, we divide these models into the following groups:

\paragraph{Traditional Methods for Crack Detection}
Here we utilize CrackForest~\cite{shi2016automatic} as a representative of traditional methods for pixel-level crack detection, since it achieves excellent performance ahead of deep learning systems.

\paragraph{DCNNs-based Edge Detection Methods}
We display some recent advanced edge detection methods, such as HED~\cite{xie2015holistically}, RCF~\cite{liu2017richer}, BDCN~\cite{he2019bi}, and BDP-Net~\cite{electronics2021likai}.

\paragraph{DCNNs-based Image Segmentation Methods for Natural Scenes}
These models includes FCN8s~\cite{long2015fully}, U-Net~\cite{ronneberger2015u}, SegNet~\cite{badrinarayanan2017segnet}, and DeepLab V3+~\cite{chen2018encoder}.

\paragraph{DCNNs-based Image Segmentation Methods for Street Scenes}
Different from the above methods, these systems, such as ENet~\cite{paszke2016enet}, ERFNet~\cite{romera2018erfnet}, BiSeNet V1~\cite{yu2018bisenet}, Fast-SCNN~\cite{poudel2019fast}, and BiSeNet V2~\cite{yu2021bisenet} tend to improve model efficiency\footnote{Note that for the model BiSeNet V2, we only use the final output for crack detection, because it is difficult to train all the side outputs and the final output at the same time.}.

\paragraph{DCNNs-based Crack Detection Methods}
They contain DeepCrack~\cite{zou2018deepcrack}, FPCNet~\cite{liu2019fpcnet}, FPHBN~\cite{yang2019feature}, CrackSeg~\cite{song2020automated}, U-CliqueNet~{\cite{Li2020Automatic}}, AM-MFF~\cite{Qu2021AMMFF}, MFF~\cite{Qu2021MFF}, and our CarNet.

Given the limited scale of crack data, DeepLab V3+ and CrackSeg are rebuilt based on ResNet-50 instead of ResNet-101.
In addition, all models employ the original output without any post-processing, such as expansion, corrosion, non-maximum suppression, and sharpening.

\begin{table*}[]
    \centering
    \scriptsize
    \caption{Experimental results on the crack databases Sun520, Rain365, and BJN260. {For convenience, we utilize red, green, and blue respectively to highlight the top three methods in terms of model accuracy or efficiency.}}
    \begin{tabular*}{\hsize}{@{}@{\extracolsep{\fill}}c|cc|cc|cc|c|c|ccc@{}}
        \hline
        \multirow{2}{*}{Methods} & \multicolumn{2}{c|}{Sun520}                        & \multicolumn{2}{c|}{Rain365}                       & \multicolumn{2}{c|}{BJN260}                        & \multirow{2}{*}{{Params}} & \multirow{2}{*}{FLOPs} & \multirow{2}{*}{FPS} \\
        \cline{2-7}
          &{ODS} &{OIS} &{ODS} &{OIS} &{ODS} & {OIS} &     &       &                      \\
        \hline
        CrackForest~\cite{shi2016automatic}    & 0.4172  & 0.4660   & 0.4898   & \textcolor{blue}{0.5277}   & 0.4550  & 0.4415  & -                          & -   & -    \\
        \hline
        HED~\cite{xie2015holistically}                     & 0.4351                  & 0.4512                  & 0.4953                  & 0.5026                  & 0.4943                  & 0.5035                  & 14.72 M                     & 47.32~G                      & 54.10 \\
        RCF~\cite{liu2017richer}                   & 0.4397                  & 0.4538                   & {0.5071} & {0.5170}                 & 0.5147                  & 0.5228                  & 14.80 M                     & 60.16~G                      & 39.63 \\
        BDCN~\cite{he2019bi}                & 0.4526                  & 0.4683                  & 0.5060                   & 0.5148 & \textcolor{blue}{0.5203}                & 0.5276                  & 16.30 M                     & 84.23~G                      & 23.97 \\
        BDP-Net~\cite{electronics2021likai}               & \textcolor{green}{0.4877} & \textcolor{green}{0.5006}
        & \textcolor{green}{0.5305} & \textcolor{green}{0.5485}               & \textcolor{green}{0.5382} &\textcolor{green}{0.5414}                & 14.80 M                     & 60.16~G                      & 37.06 \\                \hline
        FCN8s~\cite{long2015fully}                  & 0.4352                  & 0.4431                  & 0.4999                  & 0.5065                  & 0.5060                   & 0.5101                  & 134.27 M                   & 129.53~G                     & 22.45 \\
        U-Net~\cite{ronneberger2015u}              & 0.4461                  & 0.4506                  & 0.5056                  & 0.5119                  & 0.5123                  & 0.5200                    & 31.03 M                    & 128.18~G                     & 36.46 \\
        SegNet~\cite{badrinarayanan2017segnet}               & 0.4291                  & 0.4334                  & 0.4970                   & 0.5051                  & 0.4674                  & 0.4491                  & 29.44 M                    & 93.99~G                      & 40.48 \\
        DeepLab V3+~\cite{chen2018encoder}             & 0.4092                  & 0.4218                   & 0.4309                  & 0.4561                  & 0.4225                  & 0.4420                   & 40.35 M                    & 59.76~G                      & 30.19 \\
        \hline
        ENet~\cite{paszke2016enet}                      & 0.4275                  & 0.4399                  & 0.4832                  & 0.4918                  & 0.4852                  & 0.4935                  & \textbf{349.07 K}                   & 1.23~G                       & 62.20 \\
        ERFNet~\cite{romera2018erfnet}                   & 0.4303                  & 0.4407                  & 0.4837                  & 0.4918                  & 0.4787                  & 0.4822                  & 2.06 M                     & 8.63~G                       & 88.07 \\
        BiSeNet V1~\cite{yu2018bisenet}                 & 0.3941                  & 0.4190                   & 0.4693                  & 0.4834                  & 0.4381                  & 0.4583                  & 13.25 M                    & 8.74~G                       & \textcolor{green}{105.98} \\
        Fast-SCNN~\cite{poudel2019fast}                 & 0.3780                  & 0.3915                  & 0.4249                  & 0.4327                  & 0.3738                  & 0.3850                   & 1.14 M                     & \textbf{0.51~G}                        & \textcolor{red}{110.80} \\
        BiSeNet V2~\cite{yu2021bisenet}                & 0.4300                    & 0.4348                  & 0.4991                  & 0.5045                  & 0.4872                  & 0.4998                  & 3.40 M                      & 7.36~G                       & 100.19 \\
        \hline
        DeepCrack~\cite{zou2018deepcrack}               & 0.4483                  & 0.4621                 & 0.49294
                          & 0.50326                  & 0.4923                  & 0.4955                  & 29.48 M                    & 99.85~G                      & 30.70 \\
        FPCNet~\cite{liu2019fpcnet}          & \textcolor{blue}{0.4602}  & \textcolor{blue}{0.4690} & \textcolor{blue}{0.5118} & 0.5126    & 0.5010    & 0.5068  & 25.35 M  & 62.38 G  & 45.00 \\
        FPHBN~\cite{yang2019feature}           & 0.4444  & 0.4600 & 0.5073 & 0.5191    & 0.5158    & 0.5247  & 34.92 M  & 147.85 G & 28.19 \\
        CrackSeg~\cite{song2020automated}               & 0.4204                  & 0.4259                  & 0.4767                  & 0.4794                  & 0.4748                  & 0.4837                  & 53.87 M                    & 116.67~G                     & 16.93 \\
        U-CliqueNet~{\cite{Li2020Automatic}} & {0.4552} & {0.4686}                  & 0.5034                  & 0.5122                  & 0.5200   & \textcolor{blue}{0.5325}                  & 487.77 K                   & 38.52~G                      & 27.80 \\
        AM-MFF~\cite{Qu2021AMMFF}          & 0.4099  & 0.4168 & 0.4721 & 0.4749    & 0.4447    & 0.4511  & 273.50 M & 179.92 G & 12.87 \\
        MFF~\cite{Qu2021MFF}             & 0.4299  & 0.4404 & 0.4966 & 0.5046    & 0.4802    & 0.4922  & 39.73 M  & 74.80 G  & 24.80  \\
        CarNet                    & \textcolor{red}{0.5139}      & \textcolor{red}{0.5158}                &\textcolor{red}{0.5586}    & \textcolor{red}{0.5595}                  & \textcolor{red}{0.5633} & \textcolor{red}{0.5659}                  & 4.89 M                     & 11.26~G                      & \textcolor{blue}{104.37} \\
        \hline
    \end{tabular*}
    \label{Sun520_Rain365_BJN260_summary}
\end{table*}

\textbf{Experimental Results on Sun520.}
According to the results in \reftab{Sun520_Rain365_BJN260_summary}, one may find that:

{a) The proposed CarNet outperforms other state-of-the-art (SOTA) methods in terms of model performance on Sun520.}
Compared with CrackForest~\cite{shi2016automatic}, a non-deep learning method, our CarNet achieves at least 9\% and 4\% improvement in ODS and OIS, respectively.
Meanwhile, for ODS and OIS, our CarNet is respectively at least 2\% and 1\% higher than the second-ranked method BDP-Net\cite{electronics2021likai},
and 5\% and 4\% higher than the third-ranked system FPCNet~\cite{liu2019fpcnet}.
In addition, concerning model efficiency, our CarNet is about 3 and 2 times as fast as BDP-Net and FPCNet, respectively.

{Note that although other deep learning algorithms, such as BDP-Net~\cite{electronics2021likai}, DeepLab V3+~\cite{chen2018encoder}, DeepCrack~\cite{zou2018deepcrack}, FPCNet~\cite{liu2019fpcnet}, and FPHBN~\cite{yang2019feature}, also utilize multi-scale techniques, they are still inferior to ours in terms of model accuracy.
We believe this is due to the fact that in addition to multi-scale features, the encoder structure, feature up-sampling method, and feature refinement module all play a non-negligible role in model accuracy ODS and OIS, as described in Section \ref{effect of network}.}

{b) The proposed CarNet ranks third in model efficiency on Sun520.}
Note that Fast-SCNN\cite{poudel2019fast} achieves SOTA in inference speed.
Although being slightly inferior to Fast-SCNN in inference speed (slowing about 6 frames per second), our CarNet gains significant advantages by at least 13\% and 12\% respectively about ODS and OIS.
Meanwhile, under almost the same model efficiency, our CarNet achieves at least 11\% and 9\% higher in ODS and OIS respectively than the second fastest model BiSeNetV1\cite{yu2018bisenet}.

\begin{figure*}[htbp]
    \centering
    \includegraphics[width=\linewidth]{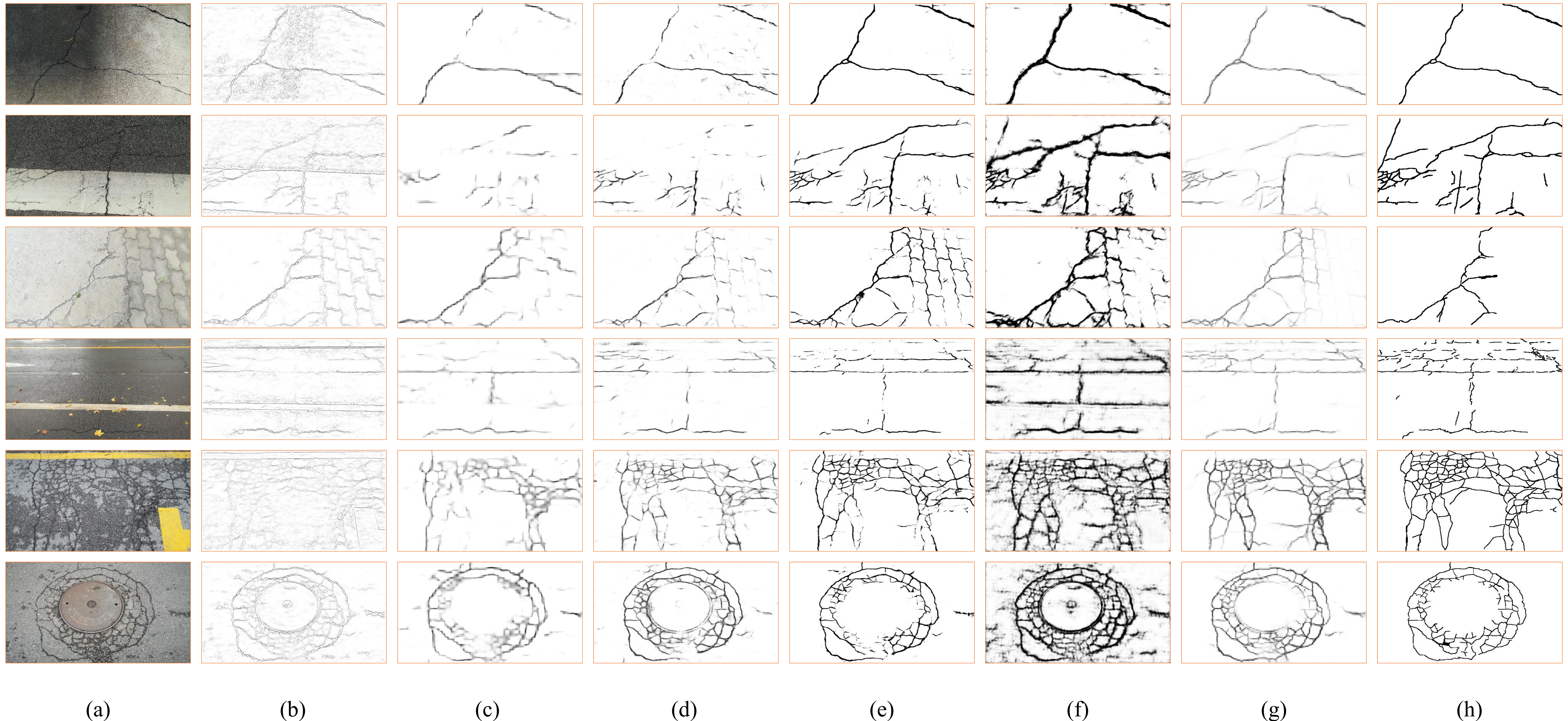}
    \caption{Visualization results of different models on Sun520 and Rain365.
    (a) and (h) contain the raw images and the corresponding ground-truth, respectively.
    (b$\sim$g) show the predicted images obtained by the models CrackForest, FastSCNN, BiSeNet V1, FPCNet, BDP-Net, and CarNet, respectively.
    Note that these results are not post-processed in this article.}
    \label{visualization on Sun520_Rain365}
\end{figure*}

In summary, the proposed CarNet achieves a good trade-off between model accuracy and efficiency.
Apart from quantitative results on Sun520, we also reveal qualitative comparisons in \reffig{visualization on Sun520_Rain365}.
Due to the limitation of space, we only show the top three methods in terms of model efficiency or accuracy.
Moreover, we also display the related results of the traditional method CrackForests~\cite{shi2016automatic}.


Compared with DCNNs-based methods including CarNet, CrackForest easily obtains false cracks.
Compared with CarNet, Fast-SCNN and BiSeNet V1 miss some true cracks while BDP-Net and FPCNet report some false cracks.
We argue that the phenomena are closely related to the designs of models.
Specifically, CrackForest utilizes the shallow model (i.e. random structured forests) to only obtain low-level features and be vulnerable to noise in the background images.
To improve inference speed, Fast-SCNN and BiSeNet V1 adopt three consecutive down-sampling with stride 2 in the initial network stage.
Then, in the decoders, they directly restore the feature maps to the model input size by eight times up-sampling.
Despite improving inference speed, they also easily miss some spatial details in the feature maps, especially for trivial tiny cracks.
{The encoders of FPCNet and BDP-Net are based on U-Net and VGG-16 respectively, leading to the same or similar number of convolutional layers in different network stages respectively.
We insist that low-level features tend to focus on local details and lead to false cracks.}
Especially, BDP-Net only utilizes three down-sampling in its encoder, and gradually merges the information from its encoder and decoder.
Due to the lack of higher-level features, this model prefers local details in crack images.

\textbf{Experimental Results on Rain365.}
According to \reftab{Sun520_Rain365_BJN260_summary}, compared with BDP-Net and FPCNet on Rain365, our CarNet achieves roughly 3\% and 5\% gains in ODS, and about 1\% and 5\% improvement in OIS, respectively.
Meanwhile, for inference speed, our model is more than 2 times faster than the above two methods.
Compared with Fast-SCNN and BiSeNet V1, despite a slight drop in inference speed, CarNet attains about 13\% and 9\% gains in ODS, and 13\% and 8\% improvement in OIS, respectively.
Besides the quantitative results, \reffig{visualization on Sun520_Rain365} displays some qualitative examples.
Compared with CarNet, Fast-SCNN and BiSeNet V1 easily miss true cracks while BDP-Net and FPCNet often obtain false cracks.

\begin{figure*}[htbp]
    \centering
    \includegraphics[width=\linewidth]{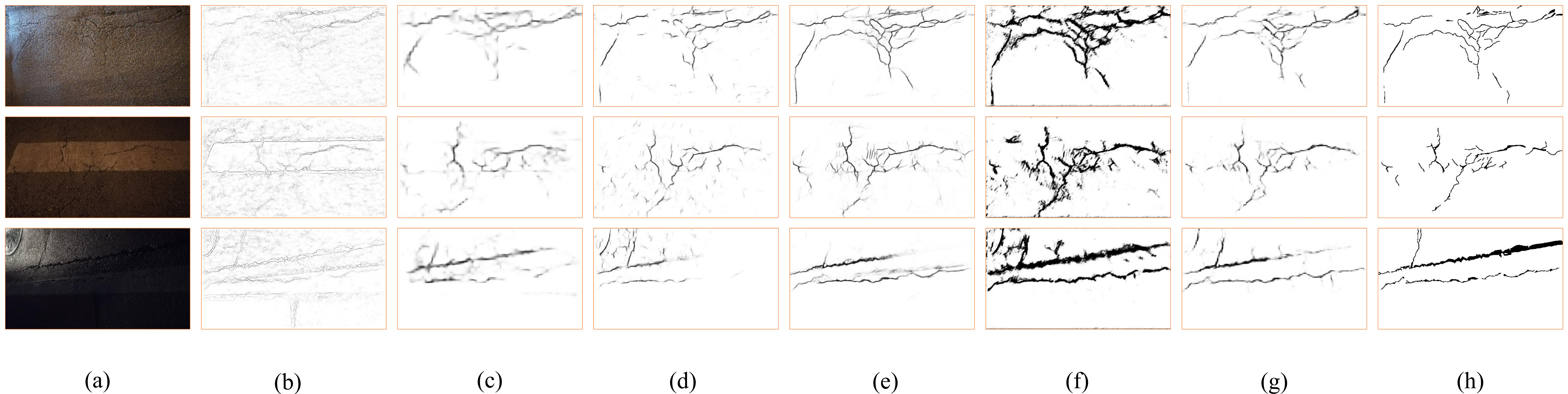}
    \caption{Visualization results of different models on BJN260.
    (a) and (g) contain the raw images and the ground-truth, respectively.
    (b$\sim$f) display the images obtained by the models CrackForest, FastSCNN, BiSeNet V1, U-CliqueNet, BDP-Net, and CarNet, respectively.
    }
    \label{visualization on BJN260}
\end{figure*}

\textbf{Experimental Results on BJN260.}
According to \reftab{Sun520_Rain365_BJN260_summary}, compared with BDP-Net and U-CliqueNet on BJN260, our CarNet obtains significant gain in inference speed while achieving about 2\% and 4\% improvement in ODS, 2\% and 3\% increase in OIS, respectively.
Compared with Fast-SCNN and BiSeNet V1, despite a slight decrease in inference speed, our CarNet is about 19\% and 12\% higher in ODS, 18\% and 10\% higher in OIS, respectively.
Besides, \reffig{visualization on BJN260} shows the prediction examples of different models on BJN260.
Note that BJN260 has a greater difference in the image content scale.
In this case, general image processing methods, especially ours, exhibit significant advantages in model accuracy compared to fast image segmentation methods.
Meanwhile, it shows the validity of our method in different scenarios.

{Besides, to broaden the application scenarios of our model, we also conduct experiments on Crack360 dataset \cite{zou2018deepcrack}.
Different from the three databases above, Crack360 includes relatively continuous cracks and have more grain-like textures in the background images, as shown in \reffig{cracks with noise}.
Please see Appendix \ref{Appendix A} for more details.
Moreover, we also display the precision-recall curves of different models on different databases via Appendix \ref{Appendix B}.}

\subsection{Discussion on Inference Efficiency}

Next, we discuss the factors that affect inference efficiency.
Through \reftab{Sun520_Rain365_BJN260_summary}, one may find that inference efficiency is not uniquely determined by time complexity.
For example, our CarNet can gain faster inference speed than ENet, even though CarNet has about 9 times FLOPs than ENet on the same database, e.g. Sun520.

According to ShuffleNet V2\cite{ma2018shufflenet}, efficient networks depend on four practical guidelines:
a) Equal input and output channel widths minimize memory access costs;
b) Too many group convolutions increase memory access cost;
c) The fragmentation of networks usually reduces model parallelism;
d) Element-wise operations should be not ignored.
Below, we utilize these principles to analyze why some models fail to achieve efficient inference.

First, DeepLab V3+~\cite{chen2018encoder} utilized enormous bottleneck layers in its encoder, leading to repeated changes in input and output channels.
ENet~\cite{paszke2016enet} employed bottleneck layers in its encoder and decoder.
Although greatly reducing the model complexity, this design also increased memory access cost, thus limiting ultimate inference efficiency.

Second, edge detection methods (e.g., HED~\cite{xie2015holistically}, RCF~\cite{liu2017richer}, BDCN~\cite{he2019bi}, and BDP-Net~\cite{electronics2021likai}) and crack detection methods (eg., FPHBN~\cite{yang2019feature}, AM-MFF~\cite{Qu2021AMMFF}, and MFF~\cite{Qu2021MFF}) employed five-branch structures and thereby reduced model parallelism.
Models like U-Net~\cite{ronneberger2015u}, SegNet~\cite{badrinarayanan2017segnet}, DeepCrack~\cite{zou2018deepcrack}, and FPCNet~\cite{liu2019fpcnet}, adopted excessive skip-layer structures and then cut down model parallelism.
Besides, U-CliqueNet~{\cite{Li2020Automatic}} frequently employed skip layers in the proposed module.
Based on DeepLab V3+, CrackSeg~\cite{song2020automated} increased new information fusion about low-level features through skip-layer.
Although improving model accuracy, this mechanism also further reduced the parallelism.

Third, for FCN8s~\cite{long2015fully}, there were too many element-wise operations in the last three $1\times1$ convolutional layers.

Besides, model efficiency is affected by other factors, such as hardware (such as GPU, CPU, and memory), platform (such as PyTorch and TensorFlow), and GPU accelerator.
For example, ERFNet~\cite{romera2018erfnet} applied asymmetric convolutions to completely replace common $3\times3$ convolutions in its architecture.
However, CUDA and cuDNN are usually designed to accelerate $3\times3$ convolutions.
Accordingly, ERFNet is not as efficient in model inference as expected.

{Moreover, we also explore other issues:
1) Does the second feature refinement still work when combined with other detection methods?
2) Can the proposed CarNet work when one uses other backbone networks as the baseline encoder?
3) Is the proposed lightweight architecture effective for other computer vision tasks?
Given the length of the article, we put them in the Appendix {\ref{Appendix C}}.
}

\section{Conclusion}\label{part5}

In this paper, we first establish two new pavement crack databases (i.e. Rain365 and Sun520) for performance evaluation, and share them to the community to facilitate related research.
Besides, we present a lightweight encoder-decoder architecture for pixel-wise crack detection.
Specifically, for the encoder network, we develop a novel olive-shaped structure by rethinking the distribution of the number of convolutional layers in different network stages.
In the decoder network, we introduce a lightweight up-sampling feature pyramid block to fuse hierarchical features from the encoder network.
Meanwhile, we propose to combine small kernel deconvolution with feature refinement module instead of bilinear interpolation and large kernel deconvolution for feature up-sampling.
Finally, extensive experiments on various crack detection databases reveal that the proposed method exceeds other SOTA systems in model performance while achieving comparable inference efficiency with the fastest approach.



{Besides, our databases and source code are also available online: \href{https://github.com/shiyanrubing/CarNet\_databases}{{https://github.com/shiyanrubing/CarNet\_databases}} and \href{https://github.com/shiyanrubing/CarNet-V1.0}{{https://github.com/shiyanrubing/CarNet-V1.0}}, respectively.}

\section{{Limitations and Future Work}}

{In our experiments, our work has proven to be quite promising. On the other hand, it does have some limitations:}

{ a) There is still some room for improvement in model performance and efficiency.
Our model is based on convolutional neural networks, which focus on image local features.
However, recent transformers, such as SegFormer~\cite{xie2021segformer} and Trans4Trans~\cite{zhang2022trans4trans}, tend to image global features.
Their combination can include local and global information of crack images and enhance the ability of feature representation, which will be one of our research directions.
Besides, our model efficiency can be further improved through some acceleration strategies, such as pruning \cite{liu2019rethinking} and quantization \cite{Yvinec_2023_WACV}, so as to facilitate crack detection on different hardware devices.}

{ b) Model generalization performance may deteriorate across databases.
As shown in \reftab{general performance on Sun520},  \reftab{general performance on Rain365}, and \reftab{general performance on BJN260},
although our method achieves better generalization performance when trained on BJN260, it is significantly inferior to FPCNet when trained on Sun520 and Rain365.
Note that the datasets Sun520 and Rain365 are taken during daytime while BJN260 is captured at nighttime.
Besides, compared with the first two datasets, BJN260 also presents richer multi-scale cracks on the images.
We think that some methods based on domain adaptation and domain generalization, such as \cite{lin2022bayesian, zhang2022towards, yao2022pcl, kang2022style}, may help to solve this problem.}

{ c) Our work is based on supervised learning, which relies heavily on annotated data.
Data annotation, especially for pixel-level tasks, often costs a lot of human and financial resources.
How to reduce the dependence on annotated data is also worthy of further exploration.
Some semi-supervised learning~\cite{kervadec2019curriculum, li2021transformation, mittal2021semi}~ and weakly supervised learning~\cite{wang2020self, zhang2020causal, chan2021a} methods are promising for crack detection in the future.
}

\begin{table}[htbp]
    \centering
    \scriptsize
    \caption{
    Performance of different models trained on Sun520 and tested on three datasets respectively.
    }
    \resizebox{0.95\columnwidth}{!}{
    \begin{tabular}{c|cc|cc|ccc}
    \hline
    \multirow{2}{*}{Models}   & \multicolumn{2}{c|}{Sun520}        & \multicolumn{2}{c|}{Rain365}       & \multicolumn{2}{c}{BJN260}    \\
    \cline{2-7}
                                                                    & ODS             & OIS             & ODS             & OIS             & ODS             & OIS
    \\
        \hline
    BDP-Net~\cite{electronics2021likai}                             & 0.4877          & 0.5006          & 0.5120          & 0.5231          & 0.3336          & 0.3388          \\
    FPCNet~\cite{liu2019fpcnet}                                     & 0.4602          & 0.469           & 0.4955          & 0.5044          & \textbf{0.4773} & \textbf{0.4774} \\
    Fast-SCNN~\cite{poudel2019fast}                                 & 0.378           & 0.3915          & 0.3940          & 0.4018          & 0.2783          & 0.2791          \\
        \hline
    CarNet                                                          & \textbf{0.5139} & \textbf{0.5158} & \textbf{0.5268} & \textbf{0.5352} & 0.3215          & 0.3256          \\
        \hline
    \end{tabular}}
        \label{general performance on Sun520}
\end{table}

\begin{table}[htbp]
    \centering
    \scriptsize
    \caption{
    Performance of different models trained on Rain365 and tested on three datasets respectively.
    }
    \resizebox{0.95\columnwidth}{!}{
    \begin{tabular}{c|cc|cc|ccc}
    \hline
    \multirow{2}{*}{Models}   & \multicolumn{2}{c|}{Sun520}        & \multicolumn{2}{c|}{Rain365}       & \multicolumn{2}{c}{BJN260}    \\
    \cline{2-7}
                                                                    & ODS             & OIS             & ODS             & OIS             & ODS             & OIS
    \\
    \hline
    BDP-Net~\cite{electronics2021likai}                             & 0.465069          & 0.483637         & 0.5305          & 0.5485          & 0.4442          & 0.4264          \\
    FPCNet~\cite{liu2019fpcnet}                                     & 0.439048          & 0.442413         & 0.5118          & 0.5126          & \textbf{0.4522} & \textbf{0.4531} \\
    Fast-SCNN~\cite{poudel2019fast}                                 & 0.362209          & 0.374696         & 0.4249          & 0.4327          & 0.3370          & 0.3426
    \\
    \hline
    CarNet                                                          & \textbf{0.505447} & \textbf{0.50804} & \textbf{0.5586} & \textbf{0.5595} & 0.3514          & 0.3426
    \\
    \hline
    \end{tabular}}
        \label{general performance on Rain365}
\end{table}

\begin{table}[htbp]
    \centering
    \scriptsize
    \caption{
    Performance of different models trained on BJN260 and tested on three datasets respectively.
    }
    \resizebox{0.95\columnwidth}{!}{
    \begin{tabular}{c|cc|cc|ccc}
    \hline
    \multirow{2}{*}{Models}    & \multicolumn{2}{c|}{Sun520}        & \multicolumn{2}{c|}{Rain365}       & \multicolumn{2}{c}{BJN260}    \\
    \cline{2-7}
    & ODS             & OIS             & ODS             & OIS             & ODS             & OIS             \\
    \hline
    BDP-Net~\cite{electronics2021likai}                             & 0.438878          & 0.453384          & 0.5110          & 0.5305          & 0.5382          & 0.5414          \\
    FPCNet~\cite{liu2019fpcnet}                                     & 0.407004          & 0.420025          & 0.4734          & 0.4821          & {0.5010}        & {0.5068} \\
    Fast-SCNN~\cite{poudel2019fast}                                 & 0.271284          & 0.292214          & 0.3377          & 0.3603          & 0.3738          & 0.3850          \\
    \hline
    CarNet                                                          & \textbf{0.476046} & \textbf{0.482241} & \textbf{0.5339} & \textbf{0.5406} & \textbf{0.5633} & \textbf{0.5659}
    \\
    \hline
    \end{tabular}}
        \label{general performance on BJN260}
\end{table}


\section*{Acknowledgment}
The authors would like to thank all the editors and anonymous reviewers for their careful reading and insightful remarks.

\bibliographystyle{IEEEtran}
\bibliography{IEEEabrv,edge_detection}


\begin{IEEEbiography}[{\includegraphics[width=1in,height=1.25in,clip,keepaspectratio]{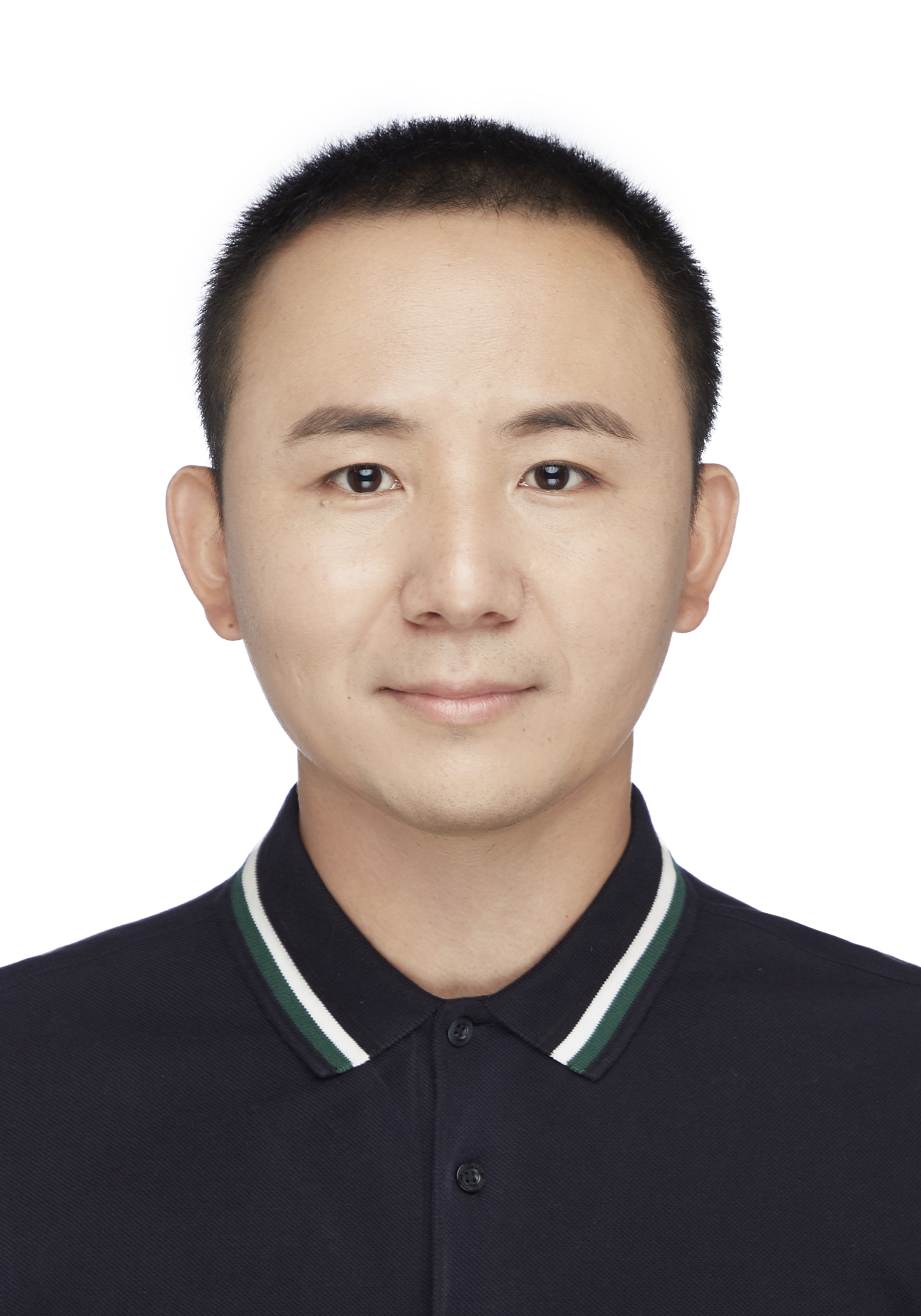}}]{Kai Li}
received the Ph.D. degrees in applied mathematics from University of Chinese Academy of Sciences, Beijing, China, in 2022.
He is currently engaged in postdoctoral research at the School of Computer Science, Peking University.
His research interests include machine learning and computer vision.
\end{IEEEbiography}

\begin{IEEEbiography}[{\includegraphics[width=1in,height=1.25in,clip,keepaspectratio]{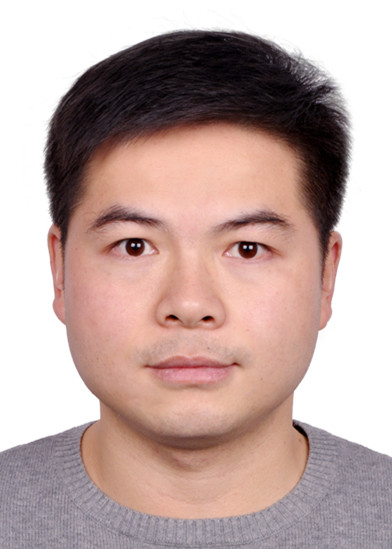}}]{Jie Yang}
received the Ph.D. degree in computer science from University of Chinese Academy of Sciences, Beijing, China, in 2021.
He is currently a research associate with the University of Chinese Academy of Sciences, Beijing, China.
His research interests include machine learning and computer vision.
\end{IEEEbiography}


\begin{IEEEbiography}[{\includegraphics[width=1in,height=1.25in,clip,keepaspectratio]{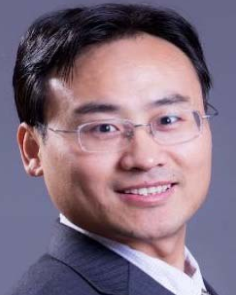}}]{Siwei Ma}
(Senior Member, IEEE) received the Ph.D. degree in computer science from the Institute of Computing Technology, Chinese Academy of Sciences, Beijing, China, in 2005.
He held a postdoctoral position with the University of Southern California, Los Angeles, CA, USA, from 2005 to 2007.
He is currently a Professor with the School of Electronics Engineering and Computer Science, Institute of Digital Media, Peking University, Beijing.
He has authored over 300 technical articles in refereed journals and proceedings in image and video coding, video processing, video streaming and transmission.
He served/serves as an Associate Editor for the IEEE Transactions on Circuits and Systems for Video Technology and the Journal of Visual Communication and Image Representation.
\end{IEEEbiography}


\begin{IEEEbiography}[{\includegraphics[width=1in,height=1.25in,clip,keepaspectratio]{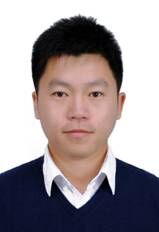}}]{Bo Wang}
received the Ph.D. degree from the University of Chinese Academy of Sciences, Beijing, in 2014.
He was also a visiting scholar with the department of computer science and engineering, Texas A\&M University, College Station, TX, USA, in 2019.
He is currently an Associate Professor with the School of Information Technology and Management, University of International Business and Economics, Beijing.
His principal research interests include statistical machine learning, optimization-based data mining, and computer vision.
\end{IEEEbiography}


\begin{IEEEbiography}[{\includegraphics[width=1in,height=1.25in,clip,keepaspectratio]{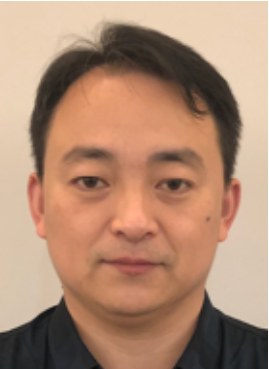}}]{Shanshe Wan}
received the Ph.D. degree in computer science from the Harbin Institute of Technology, Harbin, China, in 2014.
He held a postdoctoral position with Peking University from 2016 to 2018.
He is currently a Research Associate Professor with the School of Electronics Engineering and Computer Science, Institute of Digital Media, Peking University, Beijing.
His current research interests include video compression and image and video quality assessment.
\end{IEEEbiography}


\begin{IEEEbiography}[{\includegraphics[width=1in,height=1.25in,clip,keepaspectratio]{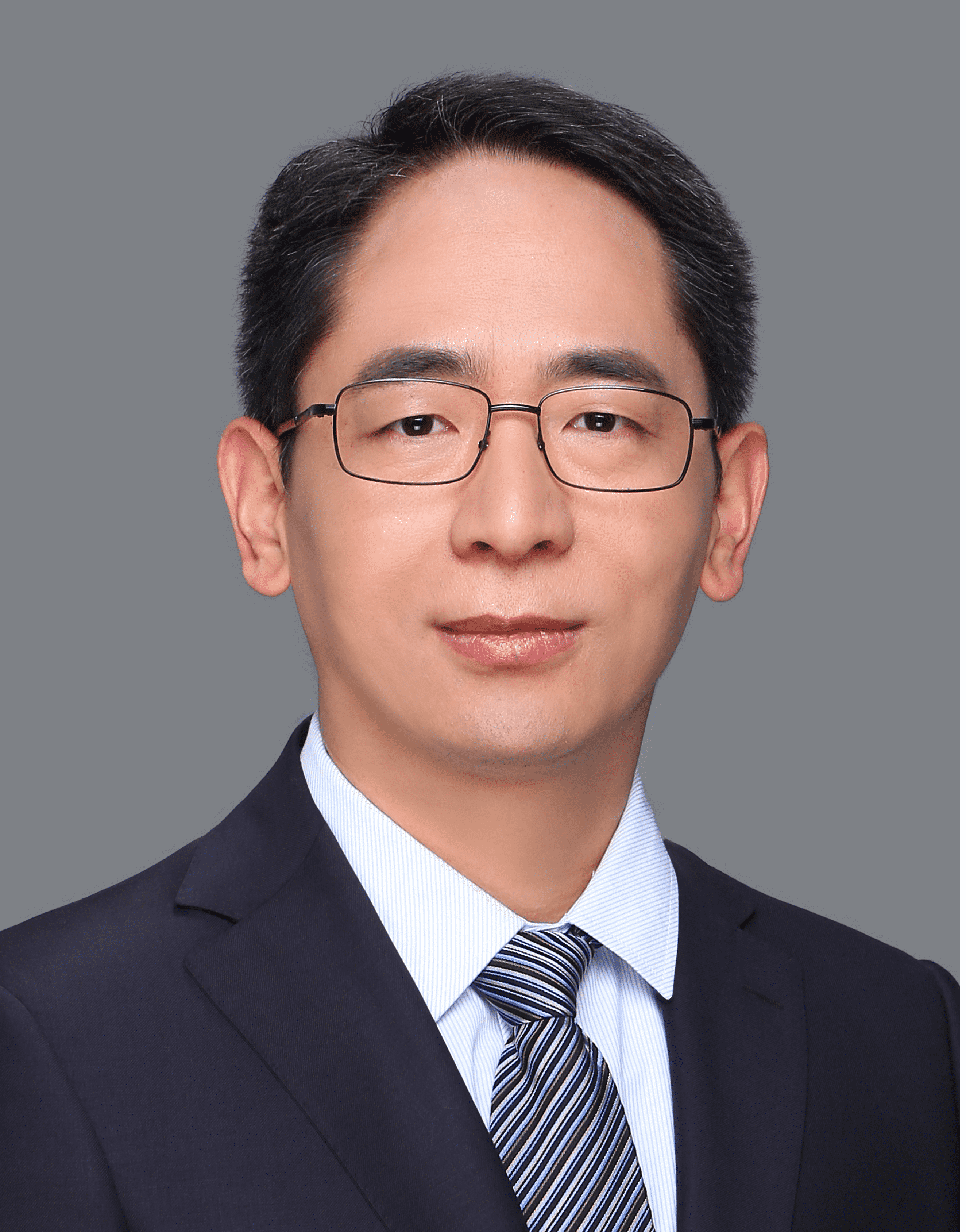}}]{Yingjie Tian}
(Member, IEEE) received the Ph.D. degree in management science and engineering from China Agricultural University, Beijing, China, in 2005.
He is currently a Professor with the Research Center on Fictitious Economy and Data Science, Chinese Academy of Sciences.
He has published four books about data mining.
His research interests include machine learning, optimization, and intelligent knowledge management.
\end{IEEEbiography}


\begin{IEEEbiography}[{\includegraphics[width=1in,height=1.25in,clip,keepaspectratio]{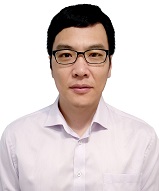}}]{Zhiquan Qi}
received the Ph.D. degrees in management science and engineering from China Agricultural University, Beijing, China, in 2011.
He is currently an Associate Professor with the Research Center on Fictitious Economy and Data Science, Chinese Academy of Sciences, Beijing.
His current research interests include deep learning and its application in weak label learning.
\end{IEEEbiography}

\clearpage


\appendix

\subsection{Experiments on Crack360} \label{Appendix A}

{To broaden the application scenarios of our model, we also utilize CrackTree260~\cite{zou2018deepcrack} and CRKWH100~\cite{zou2018deepcrack} datasets in our experiments.
Different from our Sun520, our Rain365, and BJN260 \cite{li2023fast}, they include relatively continuous cracks and have more grain-like textures in the background images, as shown in \reffig{cracks with noise}.}
Note that CrackTree260 contains 260 images with $800 \times 600$ pixels, which are cropped into $512 \times 512$ for training.
CRKWH100 embraces 100 images with $512 \times 512$ pixels for testing.
For convenience, we unify these two datasets into one database called Crack360.

According to \reftab{performance on Crack360}, compared with ERFNet and FPCNet, our CarNet gains a slightly higher ODS and OIS while achieving about 18 and 47 frames faster per second respectively in inference speed.
Compared with Fast-SCNN and BiSeNet V1, despite a drop in inference speed, CarNet obtains about 7\% and 6\% gains in ODS, 5\% and 2\% improvement in OIS, respectively.
Meanwhile, CarNet and BiSeNetV2 \cite{yu2021bisenet} have comparable model accuracy and efficiency.

\begin{table}[htbp]
    \centering
    \scriptsize
    \caption{Experimental results on the continuity-crack database Crack360.
     {For convenience, we utilize red, green, and blue respectively to highlight the top three methods in terms of model accuracy or efficiency.}}
        \resizebox{0.95\columnwidth}{!}{
    \begin{tabular*}{\hsize}{@{}@{\extracolsep{\fill}}c|cc|c|c|ccc@{}}
        \hline
        \multirow{2}{*}{Methods} & \multicolumn{2}{c|}{Crack360}
        &\multirow{2}{*}{{Params}} & \multirow{2}{*}{FLOPs} & \multirow{2}{*}{FPS} \\
        \cline{2-3}
          &{ODS} &{OIS}  &     &       &                      \\

        \hline
        CrackForest \cite{shi2016automatic}                       & 0.6604 & 0.8148     & -    & -      & -  \\
        \hline
        HED \cite{xie2015holistically}                    & 0.9352 & 0.9515       & 14.72 M           & 81.11~G         & 35.32                        \\
        RCF \cite{liu2017richer}                          & 0.9311 & 0.9491       & 14.80 M            & 102.67~G        & 25.17                        \\
        BDCN \cite{he2019bi}                              & 0.9161 & 0.9525        & 16.30 M            & 143.76~G        & 15.05                        \\
        BDP-Net \cite{electronics2021likai}               & 0.9214 & 0.9522    & 14.80 M            & 102.67~G        & 24.39                        \\
        \hline
        FCN8s \cite{long2015fully}                        & 0.9440  & 0.9630      & 134.27 M          & 189.5~G         & 13.84                        \\
        U-Net \cite{ronneberger2015u}                     & 0.9410  & 0.9529      & 31.03 M           & 218.46~G        & 22.75                        \\
        SegNet \cite{badrinarayanan2017segnet}            & 0.9407  & 0.9658      & 29.44 M           & 160.11~G        & 25.53                        \\
        DeepLab V3+ \cite{chen2018encoder}                & 0.9170  & 0.9539      & 40.35 M           & 101.16~G        & 20.25                        \\
        \hline
        ENet~\cite{paszke2016enet}                        & {0.9440}      & 0.9589     & \textbf{349.07 K} & 2.09~G          & 53.13                        \\
        ERFNet~\cite{romera2018erfnet}                    & {0.9482} & \textcolor{green}{0.9714}    & 2.06 M            & 14.72~G         & 58.71                        \\
        BiSeNet V1~\cite{yu2018bisenet}                   & 0.8939  & 0.9541    & 13.42 M           & 15.32~G         & \textcolor{green}{89.73} \\
        Fast-SCNN~\cite{poudel2019fast}                   & 0.8790  & 0.9187   & 1.14 M            & \textbf{0.87~G} & \textcolor{red}{93.52}    \\
        BiSeNet V2~\cite{yu2021bisenet}                   & 0.9407  &  {0.9699} & 3.40 M    & 12.57~G         &  \textcolor{blue}{80.24} \\
        \hline
        DeepCrack \cite{zou2018deepcrack}                 & 0.9428  & 0.9694       & 29.48 M       & 170.1~G       & 19.93            \\
        FPCNet~\cite{liu2019fpcnet}                       & \textcolor{green}{0.9512}   & 0.9682    & 25.35 M  & 106.47 G & 30.42 \\
        FPHBN~\cite{yang2019feature}                      & 0.9270   & 0.9543    & 34.92 M  & 252.32 G & 17.63 \\
        CrackSeg \cite{song2020automated}                 & 0.9434  & \textcolor{blue}{0.9700}     & 53.87 M       & 197.47~G      & 11.14                    \\
        U-CliqueNet {\cite{Li2020Automatic}}              & 0.9364  & 0.9579     & 487.77 K          & 65.57~G         & 16.41  \\
        AM-MFF~\cite{Qu2021AMMFF}                         & \textcolor{blue}{0.9497} &{0.9664}   & 273.50 M & 307.04 G & 9.41  \\
        MFF~\cite{Qu2021MFF}                              & 0.9343   & 0.9545    & 39.73 M  & 127.65 G & 16.53 \\
        CarNet                                            & \textcolor{red}{0.9536}    & \textcolor{red}{0.9724} & 4.89 M    & 19.31~G         & 77.31   \\
        \hline
    \end{tabular*}}
    \label{performance on Crack360}
\end{table}


\begin{figure*}[htbp]
    \centering
    \scriptsize
    \includegraphics[width=\linewidth]{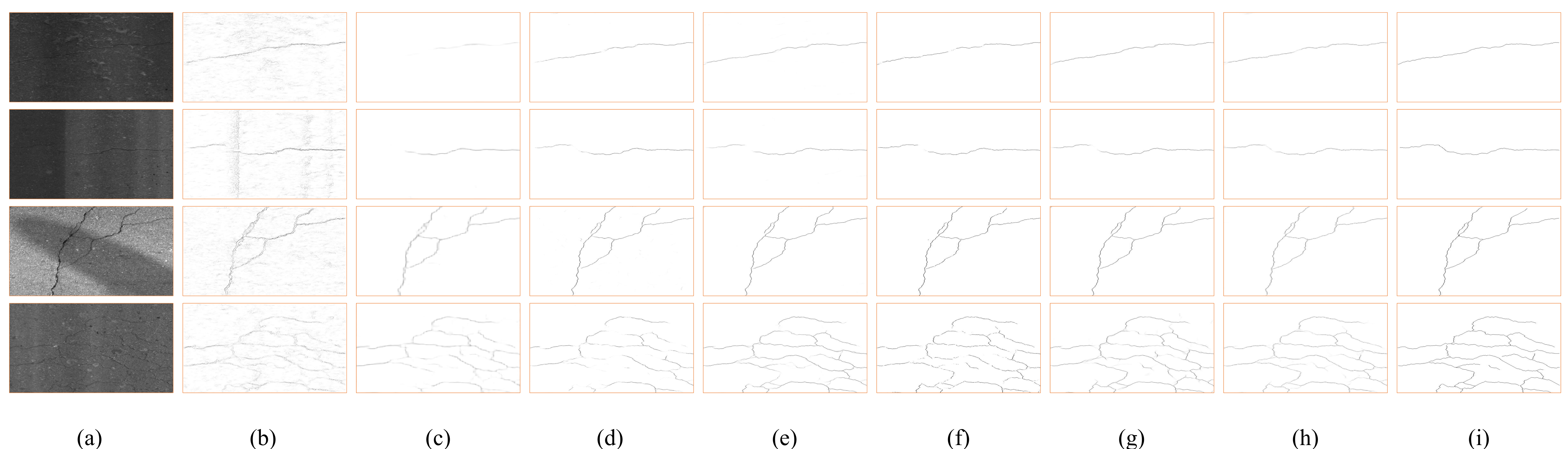}
    \caption{Visualization results of different models on Crack360.
    (a) and (i) contain the raw images and the ground-truth, respectively.
    (b$\sim$h) show the prediction images obtained by the models CrackForest, FastSCNN, BiSeNet V1, BiSeNet V2, FPCNet, ERFNet, and CarNet, respectively.}
    \label{visualization on Crack360}
\end{figure*}

On Crack360, the fast segmentation models ENet \cite{paszke2016enet}, ERFNet\cite{romera2018erfnet}, and BiSeNetV2 \cite{yu2021bisenet}, get comparable or better test accuracy than the general edge detection and image segmentation systems, which are inconsistent with the results on the three above databases.
The phenomenon is mainly due to the distribution of crack data.
Specifically, the crack images from Crack360 have few intermittent and trivial cracks which lead to minor differences in the image content scale.
To observe the crack data and the predicted results, we display some visualization examples in \reffig{visualization on Crack360}.

Besides, we also display the precision-recall curves of different models on Crack360 via Appendix \ref{Appendix B}.

\subsection{PR Curves of Different Models on Different Databases}\label{Appendix B}

\begin{figure*}
    \centering
        \scriptsize
    \subfigure[\tiny{}]{\label{PR curve_on four_datasets:a}
    \includegraphics[width=0.45\linewidth]{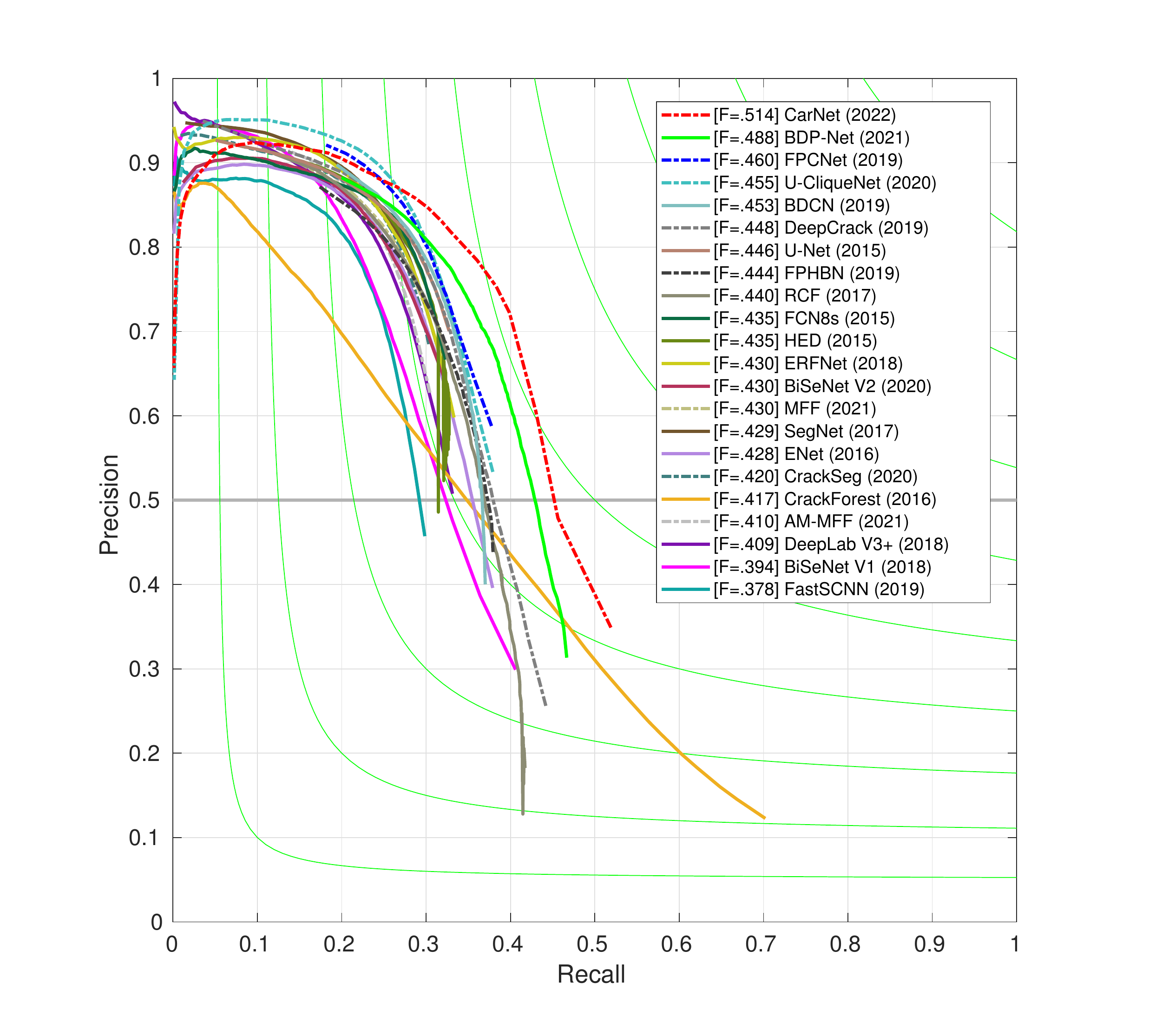}}
    \hspace{0.01\linewidth}
    \subfigure[\tiny{}]{\label{PR curve_on four_datasets:b}
    \includegraphics[width=0.45\linewidth]{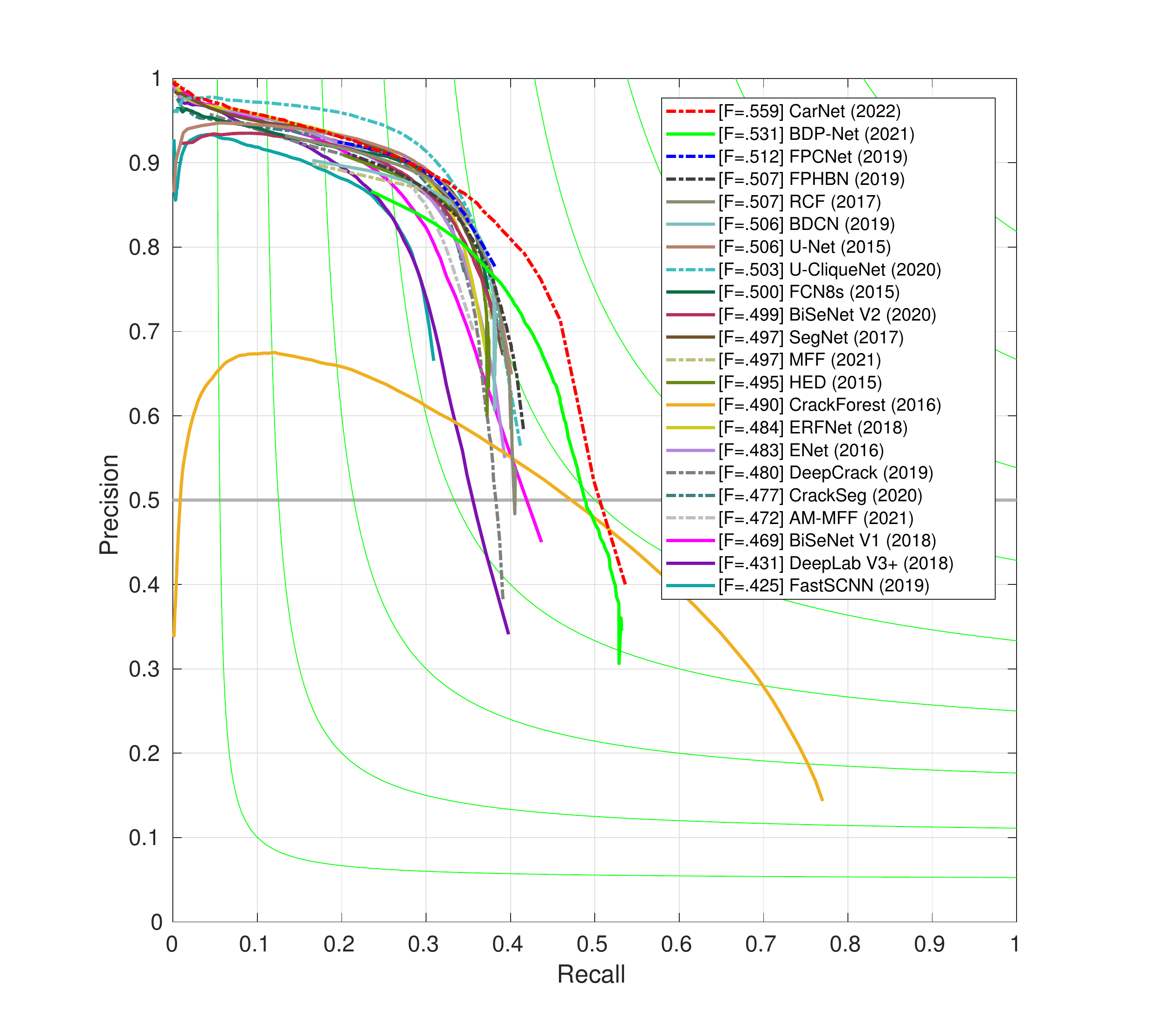}}
    \vfill
    \subfigure[\tiny{}]{\label{PR curve_on four_datasets:c}
    \includegraphics[width=0.45\linewidth]{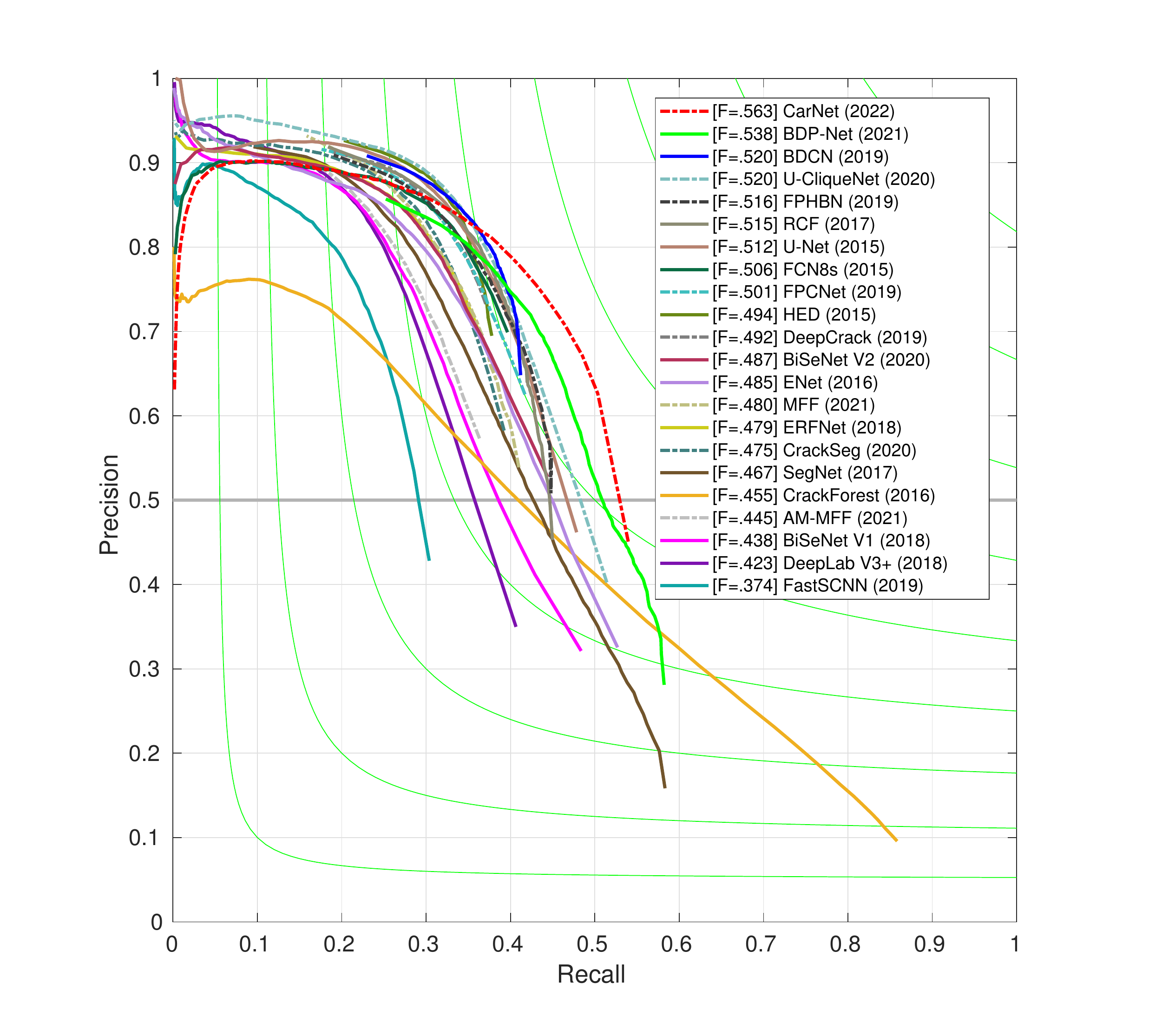}}
    \hspace{0.01\linewidth}
    \subfigure[\tiny{}]{\label{PR curve_on four_datasets:d}
    \includegraphics[width=0.45\linewidth]{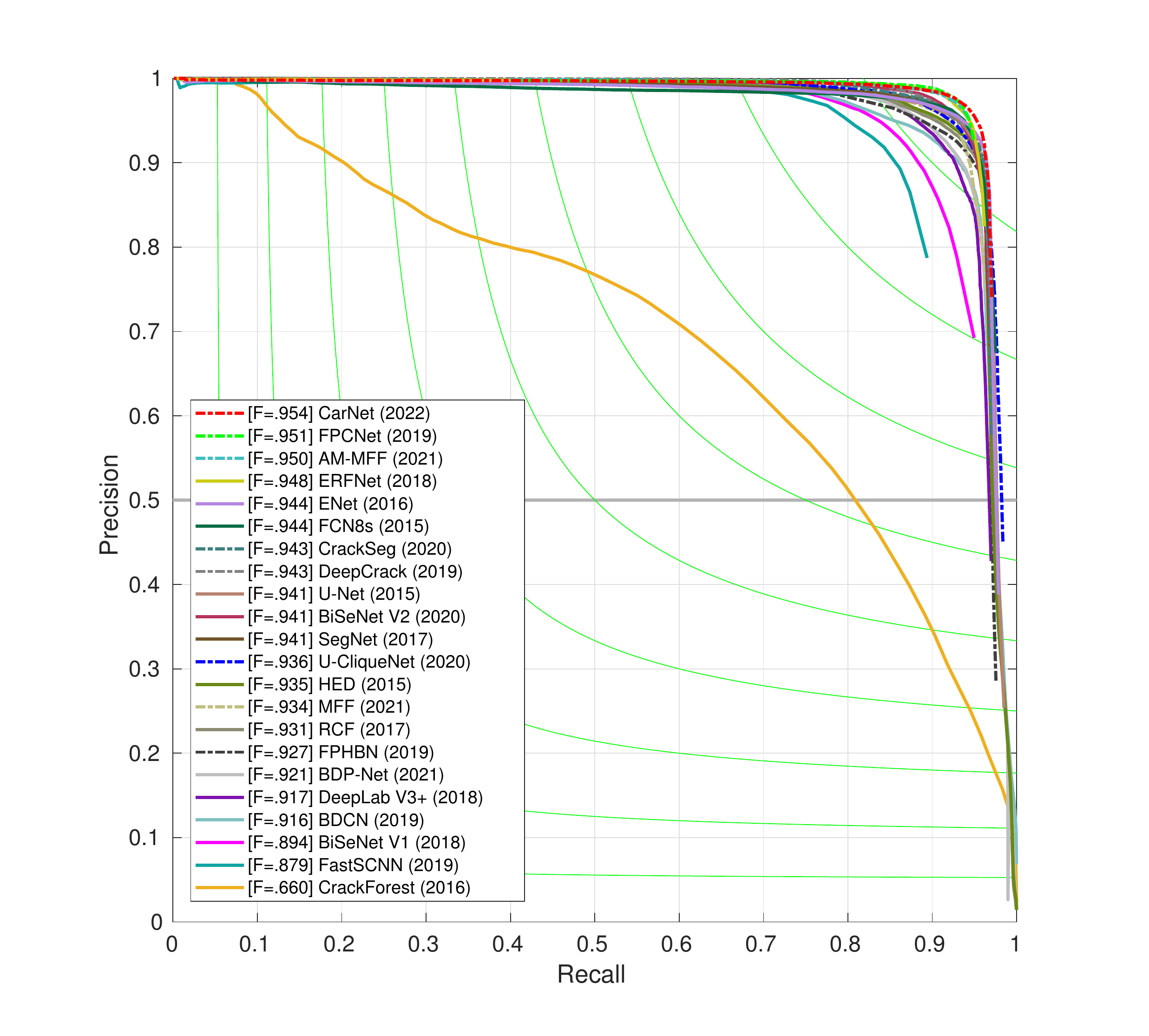}}
    \caption{Precision-Recall curves of different models on (a) Sun520, (b) Rain365, (c) BJN260,  and (d) Crack360, respectively.
    Note that the metrics precision and recall in this article are obtained by the optimal dataset scale during evaluation.}
    \label{PR curve_on four_datasets}
\end{figure*}

Here, with precision-recall curves in \reffig{PR curve_on four_datasets}, we display the precision and recall of different models on different crack databases.
Compared with other SOTA systems, our method shows some advantages in the metrics precision and recall, respectively.

\subsection{Other Discussions} \label{Appendix C}

Besides model efficiency, we also consider several other issues as follows:

1) According to \reffig{Upsampling_FR}, the second feature refinement, i.e. the one before the classifier, plays an important role in model performance.
However, far too little attention has been paid to this issue.
Then we investigate whether it still works when combined with other detection methods.

To verify its effect, we pick three methods using feature fusion (i.e. HED \cite{xie2015holistically}, DeepLab V3+ \cite{chen2018encoder}, and DeepCrack \cite{zou2018deepcrack}) and the fastest detection method (i.e. Fast-SCNN~\cite{poudel2019fast}).
Experimental results in \reftab{DCB_with_other_models} show that HED and DeepLab V3+ gain about 0.5\% and more improvement in model performance when we add the feature refinement module DCB before the classifier.
On the other hand, it does little to help with the performance of DeepCrack and Fast-SCNN.
Note that compared with the first two methods,  the latter two apply too many and too few multi-scale features, respectively.
This indicates that the feature refinement prior to the classifier has a certain scope of application.

\begin{table}[htbp]
    \centering
    \caption{Experimental results of different models combined without or with the second feature refinement module.
    These experiments are conducted on Sun520.}
    \resizebox{0.75\columnwidth}{!}{
    \begin{tabular}{c|c|ccc}
        \hline
         Models                      & ODS    & OIS \\
        \hline
         HED                         & 0.4351             & 0.4512   \\
         HED w/ FR2          & 0.4458 (\textbf{+ 1.07\%})  & 0.4604 (\textbf{+ 0.92\%}) \\
         \hline
         DeepLab V3+                 & 0.4092             & 0.4218  \\
         DeepLab V3+ w/ FR2  & 0.4192 (\textbf{+ 1\%})    & 0.4264 (\textbf{+ 0.46\%}) \\
         \hline
         DeepCrack                   & 0.4483             & 0.4621  \\
         DeepCrack w/ FR2    & 0.4508 (+ 0.15\%)  & 0.4629  (+ 0.07\%) \\
         \hline
         Fast-SCNN                   & 0.3780             & 0.3915  \\
         Fast-SCNN w/ FR2    & 0.3817 (+ 0.37\%)  & 0.3948 (+ 0.33\%)\\
         \hline
    \end{tabular}
    }
    \label{DCB_with_other_models}
\end{table}

2) Can the proposed CarNet work when using other backbone networks as the baseline encoder?

Here we employ trimmed ResNet-18 and ResNet-50 as the benchmarks.
Note that feature maps are compressed to 24 and 48 channels respectively in the multi-scale block UFPB.
The results in \reftab{ResNet-18 and ResNet-50} reveal that the proposed CarNet is still effective when based on ResNet-18 and ResNet-50, respectively.

\begin{table}[htbp]
    \centering
    \caption{Other ablation experiments on Sun520.
    Here we utilize the trimmed ResNet-18 and ResNet-50 as the encoders for the benchmark models,
    which are shown in the second and third columns respectively.
}
    \resizebox{0.98\columnwidth}{!}{
    \begin{tabular}{c|c|c|c|c|c|ccc}
        \hline
        Models & PC & ODS & OIS & Params & FLOPs & FPS \\
        \hline
        Baseline$^{\dag}$      & 2, 2, 2, 2 & 0.4379 & 0.4508 & 14.27 M         & 9.21 G          & 115.68 \\
        Baseline$^{\ddag}$       & 2, 2, 2, 2 & 0.4181 & 0.4293 & 3.90 M          & \textbf{1.84 G} & \textbf{126.38} \\
        \textbf{CarNet}         & 4, 2, 1    & \textbf{0.4546} & \textbf{0.4676}  & \textbf{2.28 M} & 5.33 G          & 108.99 \\
         \hline
        Baseline$^{\dag}$      & 3, 4, 6, 3 & 0.4449 & 0.4549 & 66.86 M         & 69.62 G         & 43.3   \\
        Baseline$^{\ddag}$       & 3, 4, 6, 3 & 0.4290 & 0.4444 & 17.91 M          & 14.13 G         & \textbf{80.589} \\
        \textbf{CarNet}   & 7, 6, 2    & \textbf{0.4626} & \textbf{0.4796} & \textbf{7.82 M}          & \textbf{11.62 G}         & 77.129 \\
         \hline
    \end{tabular}
    }
    \label{ResNet-18 and ResNet-50}
\end{table}


3) Is the proposed lightweight architecture effective for other computer vision tasks?

Here we take the edge detection task as an example.
To improve the model performance, we also use the pre-trained weights to initialize CarNet-762.
Here CarNet-762 refers to CarNet with the residual block combinations [7, 6, 2] in the second, third, and fourth stages of the encoder network.
To this end, we first construct a classification network by adding a global pooling layer and a fully connected layer behind our olive-type encoder.
Second, we utilize cross-entropy to train the classification model on ImageNet 1K database~\cite{2009ImageNet}.
Besides, we employ the hyper-parameter as follow: mini-batch size (128), initial learning rate ($1\times 10^{-2}$), momentum (0.9), weight decay ($5\times 10^{-5}$), training epochs (60). Concerning the learning rate decay, we adopt the cosine annealing scheduler.

In the following, we first introduce edge detection databases and detailed implementation, then compare the performance of our CarNet and other state-of-the-art models.

We conduct experiments on three public datasets, i.e., BSDS500 \cite{Arbel2011Contour}, NYUDv2 \cite{Silberman2012Indoor}, and Multicue \cite{MA}.
BSDS500 contains 200, 100, and~200 images for training, validation, and~testing, respectively.
NYUDv2 consists of 381, 414, and~654 images for training, validation, and~testing, respectively.
Multicue contains 80 and 20 images for training and testing, respectively.
Besides, in the above three databases, the sizes of the test images are $321\times481$, $425\times560$, and $720\times1280$, respectively.

In the experiments, for every database, we apply the training and validation datasets (if necessary) for fine-tuning, and the test dataset for evaluation.
Regarding data augmentation, we utilize the same strategy as \cite{liu2017richer}.
During training, for BSDS500 and NYUDv2, we employ the full resolution images as the model input.
As the original images on Multicue give high resolutions, we adopt the randomly cropped $500\times500$ image patches as the model input.

The mini-batch size is set to 4, 2, and 2 for BSDS500, NYUDv2, and Multicue, respectively.
The Adam optimizer \cite{kingma2015adam} is adopted to update network parameters.
The learning rate is initially set to $1\times10^{-4}$ and then divided by 10 after 15 K iterations.
The weight decay is set to $2\times 10^{-4}$. We train CarNet with 21 K, 18 K, and 1.5 K iterations on BSDS500, NYUDv2, and Multicue, respectively.
Considering that the edge detection data is generally marked by multiple people, and there are inconsistent markings, we employ weighted cross-entropy in \cite{liu2017richer} for model training and then process the predicted images through non-maximum suppression.

Through \reftab{performance on BSDS500}, \reftab{performance on NYUDv2}, and \reftab{performance on Multicue}, we present the test accuracy, space and time complexity, and inference speed of different models on BSDS500, NYUDv2, and Multicue, respectively.
Note that we reproduce the other four models to obtain the corresponding experimental results.
Moreover, the metric ODS and OIS are obtained by the single-scale test.

\begin{figure*}[htbp]
    \centering
    \subfigure[\tiny{}]{
    \centering
    \includegraphics[width=0.315\linewidth]{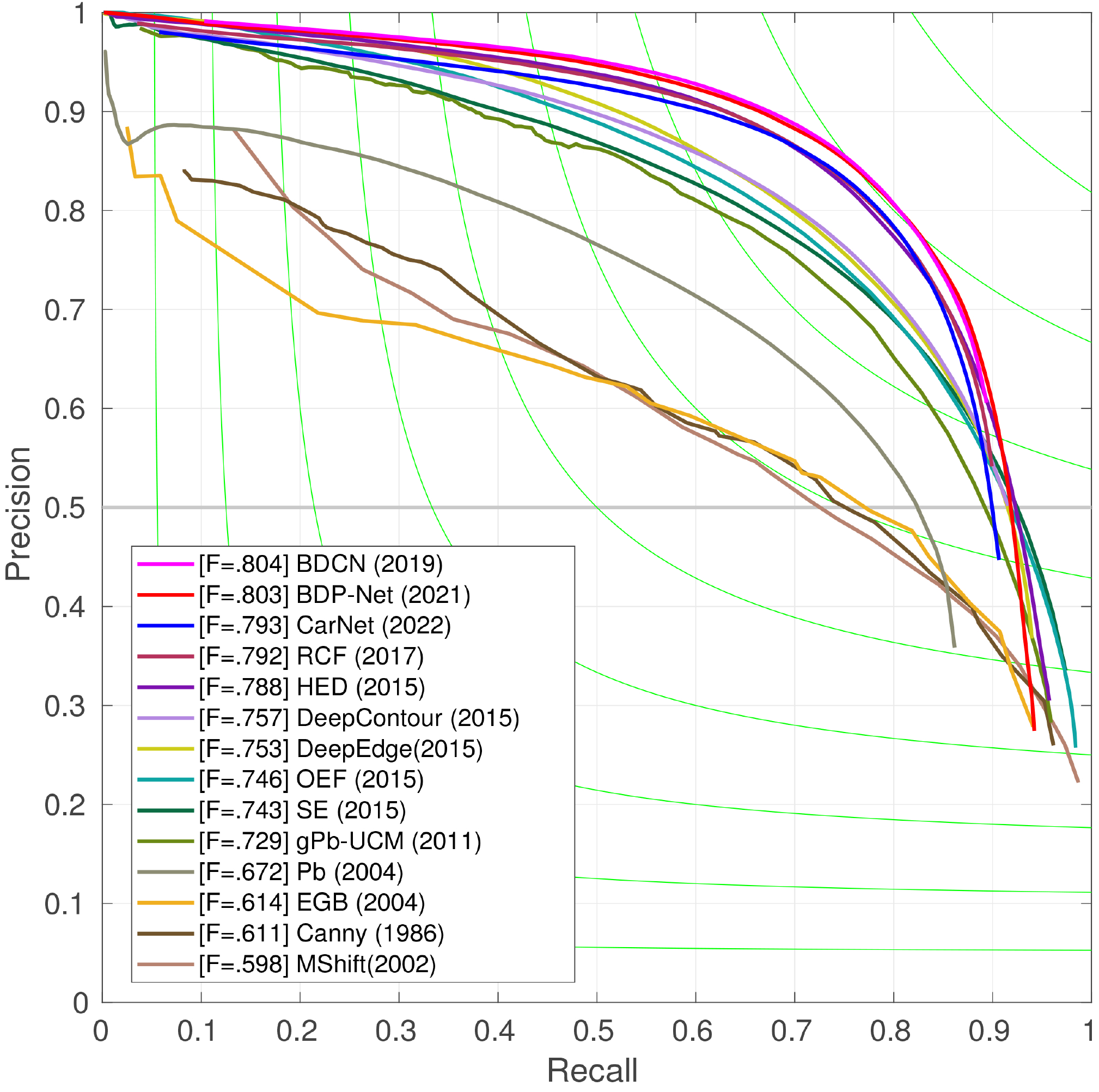}
    \label{bsds500_pr} %
    }
    \subfigure[\tiny{}]{
    \centering
   \includegraphics[width=0.315\linewidth]{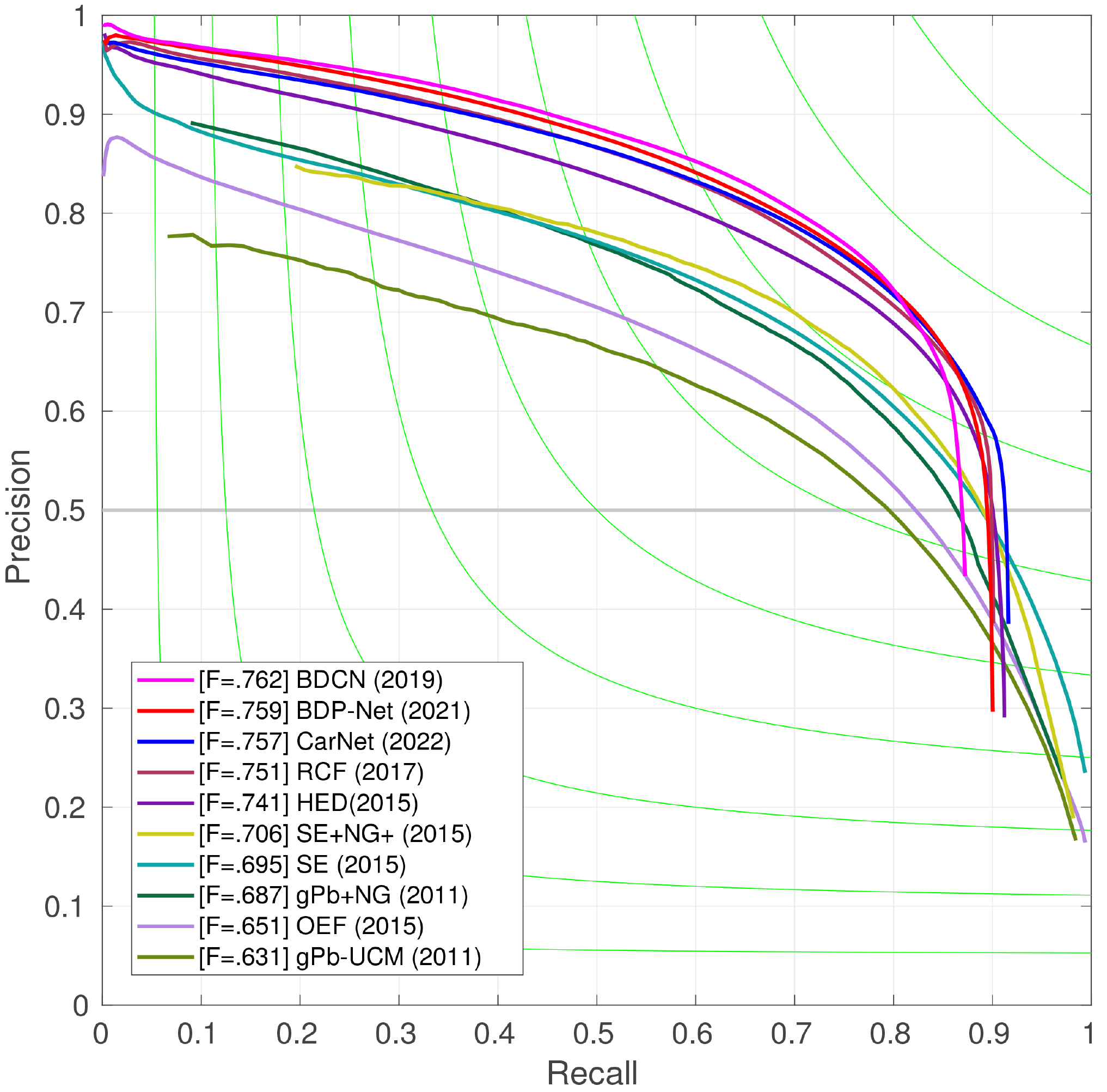}
   \label{nyud_pr} %
    }
    \subfigure[\tiny{}]{
    \centering
    \includegraphics[width=0.315\linewidth]{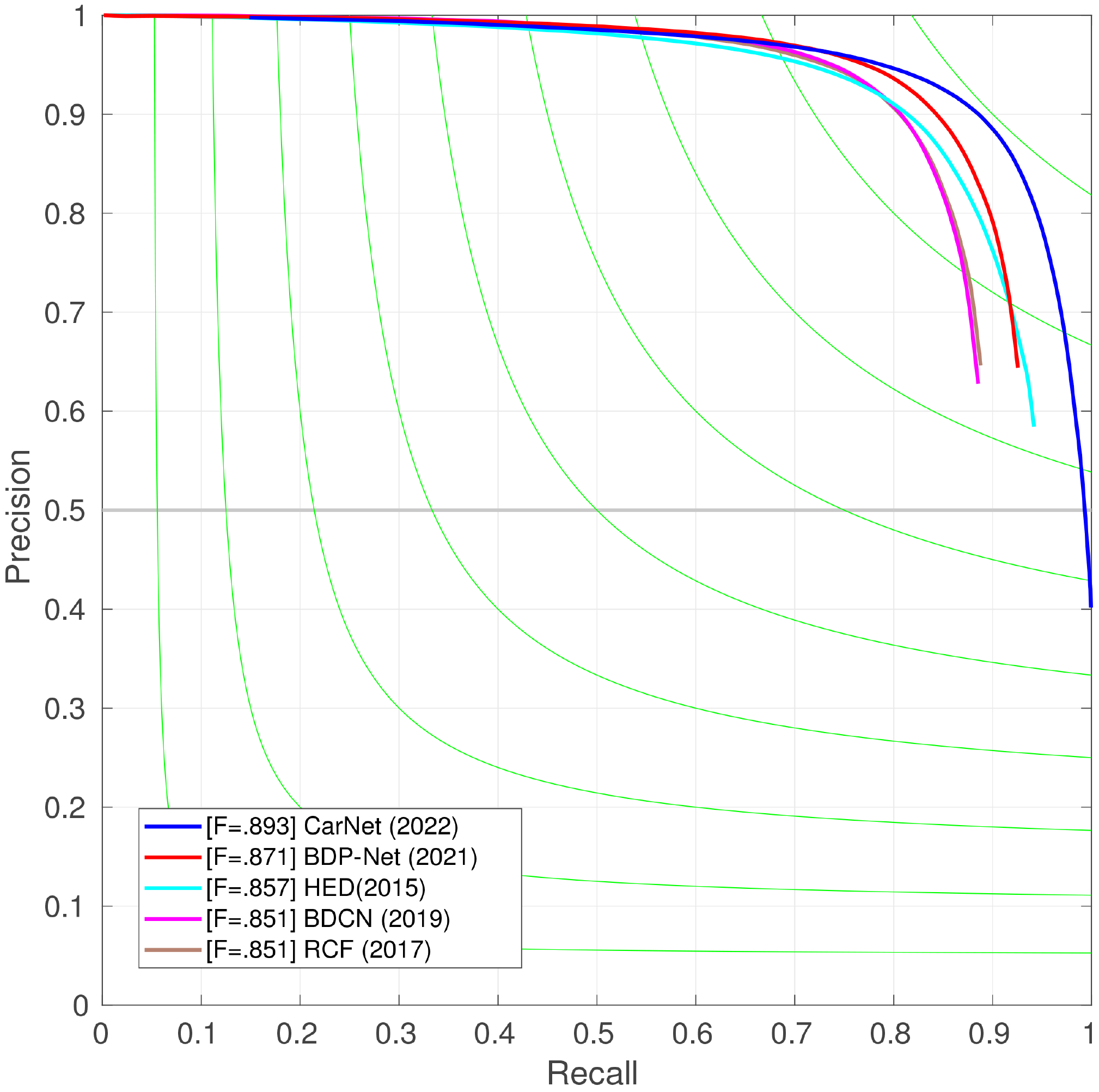}
    \label{multicue_pr}
    }
    \caption{The precision-recall curves of various algorithms on (a) BSDS500, (b) NYUDv2, and (c) Multicue, respectively.
    }
    \label{PR_curve_edge_detection}
\end{figure*}

On BSDS500, our CarNet achieves nearly two times as fast as HED in terms of inference speed while obtaining minor improvements in ODS and OIS, respectively.
Meanwhile, our model achieves two times faster than RCF in inference speed while gaining equivalent test accuracy about ODS and OIS separately.
Compared with BDCN and BDP-Net, despite a slight drop in ODS and OIS, our approach achieves about three and two times faster inference speed, respectively.
Besides, we present the precision-recall curves of different models on BSDS500, as shown in \reffig{bsds500_pr}.

\begin{table}[htbp]
    \centering
    \scriptsize
    \caption{The performance comparison of different models on BSDS500.}
    \resizebox{0.95\columnwidth}{!}{
    \begin{tabular}{c|c|c|c|c|ccc}
        \hline
        {Methods}       &{ODS}    &{OIS}      &{Params}   &{FLOPs}      & FPS         \\
        \hline
        Human                                       & 0.803 & 0.803   &-   &-  &-   \\
        HED \cite{xie2015holistically}              & 0.788 & 0.804 & 14.72 M  & 48.72~G  & 51.47  \\
        RCF \cite{liu2017richer}                    & 0.792 & 0.809 & 14.8 M   & 61.89~G  & 37.37  \\
        BDCN \cite{he2019bi}                        & \textbf{0.804} & {0.82}  & 16.3 M   & 86.24~G   & 23.23  \\
        BDP-Net \cite{electronics2021likai}         & 0.803 & \textbf{0.822} & 14.81 M   & 61.92~G  & 36.7  \\
        \hline
        CarNet                                      & 0.793 & 0.807 & \textbf{4.89 M}    & \textbf{11.66~G}  &\textbf{99.6}    \\
        \hline
    \end{tabular}}
    \label{performance on BSDS500}
\end{table}

On NYUDv2, compared with other systems, our method still has a significant advantage in inference speed.
Meanwhile, our method gains comparable test accuracy with BDCN, the state-of-the-art (SOTA) systems.
In addition, we also show the precision-recall curves of different algorithms in \reffig{nyud_pr}.

\begin{table}[htbp]
    \centering
    \scriptsize
    \caption{The performance comparison of different models on NYUDv2.}
    \resizebox{0.95\columnwidth}{!}{
    \begin{tabular}{c|c|c|c|c|ccc}
        \hline
        {Methods}       &{ODS}    &{OIS}      &{Params}   &{FLOPs}      & FPS         \\
        \hline
        HED \cite{xie2015holistically}              & 0.741 & 0.757 & 14.72 M  &73.99~G	& 36.35  \\
        RCF \cite{liu2017richer}                    & 0.751 & 0.77 & 14.8 M   & 94.42~G   & 28.38  \\
        BDCN \cite{he2019bi}                        & \textbf{0.762} & \textbf{0.776}  & 16.3 M   & 131.84~G  & 16.71   \\
        BDP-Net \cite{electronics2021likai}         & 0.759 &\textbf{0.776} & 14.81 M   & 94.46~G  & 27.52   \\
        \hline
        CarNet                                      & 0.757 &0.772 & \textbf{4.89 M}    & \textbf{17.6~G}  & \textbf{82.1}    \\
        \hline
    \end{tabular}}
    \label{performance on NYUDv2}
\end{table}

On Multicue, our CarNet sets new SOTA in test accuracy and inference speed, respectively.
Moreover, we display the precision-recall curves of different algorithms in \reffig{multicue_pr}.

\begin{table}[htbp]
    \centering
    \scriptsize
    \caption{The performance comparison of different models on Multicue.}
    \resizebox{0.95\columnwidth}{!}{
    \begin{tabular}{c|c|c|c|c|ccc}
        \hline
        {Methods}       &{ODS}    &{OIS}      &{Params}   &{FLOPs}      & FPS         \\
        \hline
        Human                                       & 0.75 & -   &-   &-  &-   \\
        HED \cite{xie2015holistically}              & 0.857 & 0.867 & 14.72 M  & 283.9~G	& 8.65   \\
        RCF \cite{liu2017richer}                    & 0.851 & 0.857 & 14.8 M   & 363.4~G	& 6.68  \\
        BDCN \cite{he2019bi}                        & 0.851 & 0.858  & 16.3 M   & 507.97~G	& 4.03  \\
        BDP-Net \cite{electronics2021likai}         &{0.871} &{0.877} & 14.81 M   & 363.56~G	& 6.51  \\
        \hline
        CarNet                                      & \textbf{0.893}	& \textbf{0.9} & \textbf{4.89 M}    & \textbf{67.35~G}	& \textbf{18.06}    \\
        \hline
    \end{tabular}}
    \label{performance on Multicue}
\end{table}

\begin{figure*}[htbp]
    \centering
    \includegraphics[width=\linewidth]{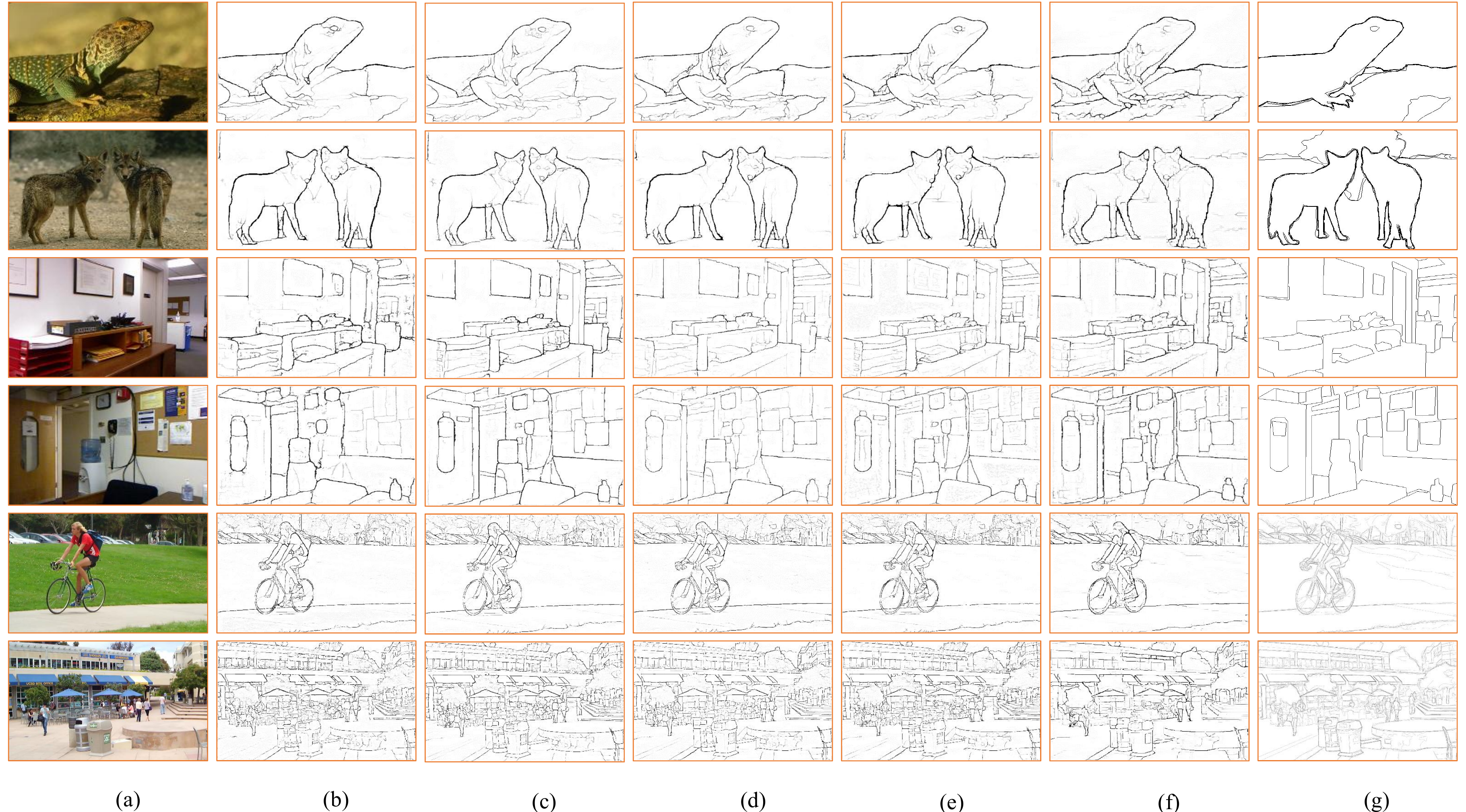}
    \caption{Visualization examples of different models.
    Here we randomly select two images from BSDS500, NYUD V2, and Multicue, and then arrange them in order from top to bottom.
    (a) and (f) contain the raw images and the ground-truth, respectively.
    (b$\sim$e) refer to the results obtained by the models HED, RCF, BDCN, BDP-Net, and CarNet, respectively.
    }
    \label{vision about edge} %
\end{figure*}

Furthermore, we also show qualitative results of different models in \reffig{vision about edge}.
The visualization examples indicate that their differences in predicted images are quite small.

In summary, our CarNet can also achieve a good trade-off between efficiency and accuracy on the edge detection task.

\end{document}